\documentclass[10pt,twocolumn,letterpaper]{article}

\usepackage{multirow}
\usepackage{tabularx}
\usepackage{array,booktabs}
\usepackage{ragged2e}
\usepackage{tabu}
\usepackage{pgfplots,pgfplotstable,filecontents}
\usepackage[space]{grffile}

\usepackage{tikz}
\usepackage{verbatim}
\usepackage[english]{babel}
\usepackage{xcolor}
\usepackage{blindtext}
\usepackage[export]{adjustbox}

\usetikzlibrary{calc,shadings}

\usepackage{pgfplots}

\def\addlegendimage{\csname pgfplots@addlegendimage\endcsname}

\pgfkeys{/pgfplots/number in legend/.style={%
        /pgfplots/legend image code/.code={%
            \node at (0.295,-0.0225){#1};
        },%
    },
}

\usepackage{caption}
\usepackage{subcaption}

\usepackage{cvpr}
\usepackage{times}
\usepackage{epsfig}
\usepackage{graphicx}
\usepackage{amsmath}
\usepackage{amssymb}
\usepackage{enumerate}

\usepackage[pagebackref=true,breaklinks=true,letterpaper=true,colorlinks,bookmarks=false]{hyperref}

\cvprfinalcopy 

\ifcvprfinal\fi
\begin{document}

\title{Seven ways to improve example-based single image super resolution}

\author{Radu Timofte\\
Computer Vision Lab\\
D-ITET, ETH Zurich\\
{\tt\small timofter@vision.ee.ethz.ch}
\and
Rasmus Rothe\\
Computer Vision Lab\\
D-ITET, ETH Zurich\\
{\tt\small rrothe@vision.ee.ethz.ch}
\and
Luc Van Gool\\
ESAT, KU Leuven\\
D-ITET, ETH Zurich\\
{\tt\small vangool@vision.ee.ethz.ch}
}
\maketitle

\begin{abstract}
In this paper we present seven techniques that everybody should know to improve example-based single image super resolution (SR): 1) augmentation of data, 2) use of large dictionaries with efficient search structures, 3) cascading, 4) image self-similarities, 5) back projection refinement, 6) enhanced prediction by consistency check, and 7) context reasoning.

We validate our seven techniques on standard SR benchmarks (\ie Set5, Set14, B100) and methods (\ie A+, SRCNN, ANR, Zeyde, Yang) and achieve substantial improvements.
The techniques are widely applicable and require no changes or only minor adjustments of the SR methods. 

Moreover, our Improved A+ (IA) method sets new state-of-the-art results outperforming A+ by up to 0.9dB on average PSNR whilst maintaining a low time complexity.
\end{abstract}

\section{Introduction}

Single image super-resolution (SR) aims at reconstructing a high-resolution (HR) image by restoring the high frequencies details from a single low-resolution (LR) image.
SR is heavily ill-posed since multiple HR patches could correspond to the same LR image patch.
To address this problem, the SR literature proposes interpolation-based methods~\cite{Thevenaz-BOOK-2000}, reconstruction-based methods~\cite{Chang-CVPR-2004,Glasner-ICCV-2009,Protter-TIP-2009,Yang-TIP-2010,Zeyde-CS-2012}, and learning-based methods~\cite{Kim-PAMI-2010,Dong-TIP-2011,Timofte-ICCV-2013,Timofte-ACCV-2014,Dong-ECCV-2014,Zhang-TIP-2015,Dai-EG-2015}.

The example-based SR~\cite{Freeman-CGA-2002} uses prior knowledge under the form of corresponding pairs of LR-HR image patches extracted internally from the input LR image or from external images. Most recent methods fit into this category. 

\begin{figure}[t!]
\includegraphics[width=\columnwidth]{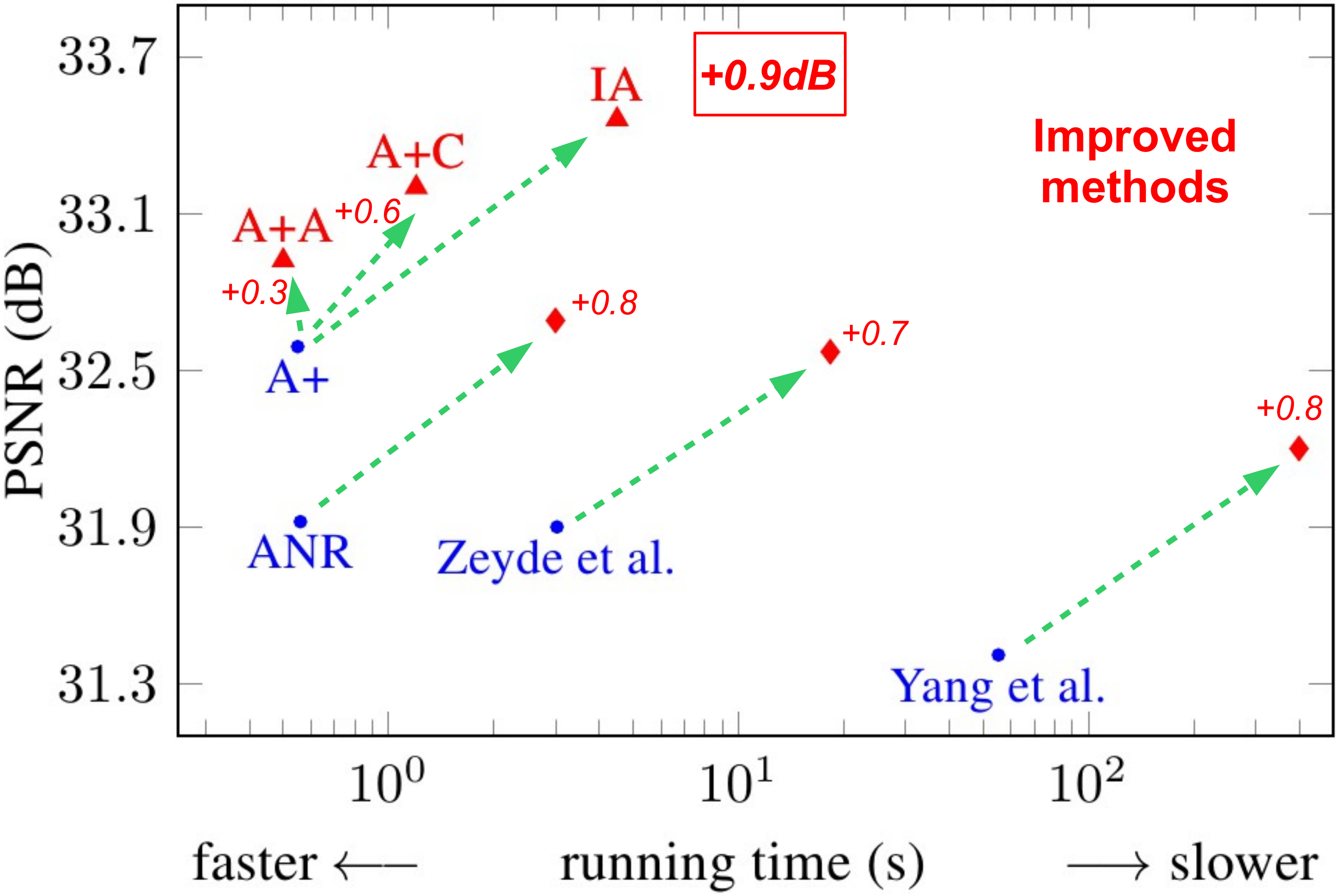}
\caption{We largely improve ({\color{red} red}) over the original example-based single image super-resolution methods ({\color{blue} blue}), \ie our IA method is 0.9dB better than A+~\cite{Timofte-ACCV-2014} and 2dB better than Yang~\etal~\cite{Yang-CVPR-2008}. Results reported on Set5, $\times3$. Details in Section~\ref{ssc:generality}.} 
\label{fig:teaser}
\vspace{-0.5cm}
\end{figure}

In this paper we present seven ways to improve example-based SR.
We apply them to the major recent methods: the Adjusted Anchored Neighborhood Regression (A+) method introduced recently by Timofte~\etal~\cite{Timofte-ACCV-2014}, the prior Anchored Neighborhood Regression (ANR) method by the same authors~\cite{Timofte-ICCV-2013}, the efficient K-SVD/OMP method of Zeyde~\etal~\cite{Zeyde-CS-2012}, the sparse coding method of Yang~\etal~\cite{Yang-CVPR-2008}, and the convolutional neural network method (SRCNN) of Dong~\etal~\cite{Dong-ECCV-2014}. We achieve consistently significant improvements on standard benchmarks. 
Also, we combine the techniques to derive our Improved A+ (IA) method.
Fig.~\ref{fig:teaser} shows a comparison of the large relative improvements when starting from the A+, ANR, Zeyde, or Yang methods on Set5 test images for magnification factor $\times3$. Zeyde is improved by 0.7dB in Peak Signal to Noise Ratio (PSNR), Yang and ANR by 0.8dB, and A+ by 0.9dB. Also, in Fig.~\ref{fig:seven} we draw a summary of improvements for A+ in relation to our proposed Improved A+ (IA) method.

The remainder of the paper is structured as follows.
First, in Section~\ref{sec:framework} we describe the framework that we use in all our experiments and briefly review the anchored regression baseline - the A+ method~\cite{Timofte-ACCV-2014}. Then in Section~\ref{sec:methods} we present the seven ways to improve SR and introduce our Improved A+ (IA) method. In Section~\ref{sec:discussion} we discuss the generality of the proposed techniques and the results, to then draw the conclusions in Section~\ref{sec:conclusion}.


\section{General framework}
\label{sec:framework}
We adopt the framework of~\cite{Timofte-ICCV-2013,Timofte-ACCV-2014} for developing our methods and running the experiments. 
As in those papers, we use 91 training images proposed by~\cite{Yang-CVPR-2008}, and work in the YCbCr color space on the luminance component while the chroma components are bicubically interpolated.
For a given magnification factor, these HR images are (bicubically) downscaled to the corresponding LR images. The magnification factor is fixed to $\times3$ when comparing the 7 techniques. 
The LR and their corresponding HR images are then used for training example-based super-resolution methods such as A+~\cite{Timofte-ACCV-2014}, ANR~\cite{Timofte-ICCV-2013},
or Zeyde~\cite{Zeyde-CS-2012}. For quantitative (PSNR) and qualitative evaluation 3 datasets Set5, Set14, and B100 are used as in~\cite{Timofte-ACCV-2014}.
In the next section we first describe the employed datasets, then the methods we use or compare with, to finally briefly review the A+~\cite{Timofte-ACCV-2014} baseline method.

\subsection{Datasets}
\label{ssc:datasets}
We use the same standard benchmarks and datasets as used in~\cite{Timofte-ACCV-2014} for introducing A+, and in~\cite{Yang-CVPR-2008, Zeyde-CS-2012, Timofte-ICCV-2013, Perez-ACCV-2014, Dong-ECCV-2014, Schulter-CVPR-2015, Dong-PAMI-2015} among others.

\begin{figure}
\centering
\scriptsize
\setlength{\tabcolsep}{1pt}     
\renewcommand{\arraystretch}{1} 
\resizebox{\linewidth}{!}{
\begin{tabular}{cccc}
\fcolorbox{red}{yellow}{\includegraphics[width=0.22\linewidth, height=0.22\linewidth]{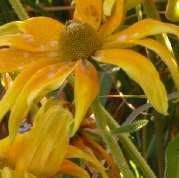}}&
\includegraphics[width=0.22\linewidth, height=0.22\linewidth]{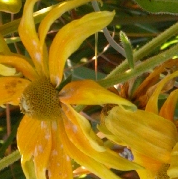}&
\includegraphics[width=0.22\linewidth, height=0.22\linewidth]{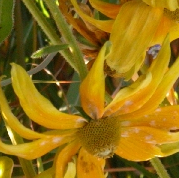}&
\includegraphics[width=0.22\linewidth, height=0.22\linewidth]{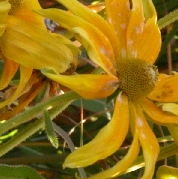}\\
original & rotated $90^{\circ}$ &rotated $180^{\circ}$ &rotated $270^{\circ}$\\
\includegraphics[width=0.22\linewidth, height=0.22\linewidth]{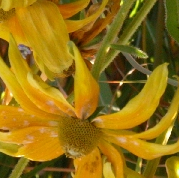}&
\includegraphics[width=0.22\linewidth, height=0.22\linewidth]{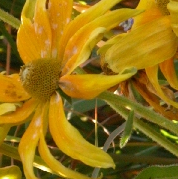}&
\includegraphics[width=0.22\linewidth, height=0.22\linewidth]{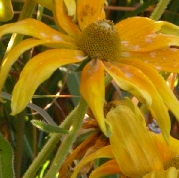}&
\includegraphics[width=0.22\linewidth, height=0.22\linewidth]{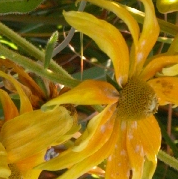}\\
flipped & $90^{\circ}$ \& flipped & $180^{\circ}$ \& flipped & $270^{\circ}$ \& flipped \\
\end{tabular}
}
\caption{Augmentation of training images by rotation and flip.}
\label{fig:augmentation}
\end{figure}

\noindent\textbf{Train91 } is a training set of 91 RGB color bitmap images as proposed by Yang~\etal~\cite{Yang-CVPR-2008}. Train91 contains mainly small sized flower images. The average image size is only $\sim6,500$ pixels.
Fig.~\ref{fig:augmentation} shows one of the training images.

\noindent\textbf{Set5 } is used for reporting results. It contains five popular images: one medium size image (`baby', $512\times512$) and four smaller ones (`bird', `butterfly',`head', `women').
They were used in~\cite{Bevilacqua-BMVC-2012} and adopted under the name `Set5' in~\cite{Timofte-ICCV-2013}.

\noindent\textbf{Set14 } is a larger, more diverse set than Set5. It contains 14 commonly used bitmap images for reporting image processing results. The images in Set14 are larger on average than those in Set5. This selection of 14 images was proposed by Zeyde~\etal~\cite{Zeyde-CS-2012}.

\noindent\textbf{B100 } is the testing set of 100 images from the Berkeley Segmentation Dataset~\cite{Martin-ICCV-2001}. The images cover a large variety of real-life scenes and all have the same size of $481\times321$ pixels. We use them for testing as in~\cite{Timofte-ACCV-2014}.

\begin{figure*}[th!]
\centering
{
\includegraphics[height=1.2cm]{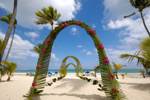}
\includegraphics[height=1.2cm]{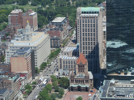}
\includegraphics[height=1.2cm]{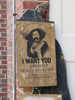}
\includegraphics[height=1.2cm]{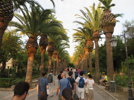}
\includegraphics[height=1.2cm]{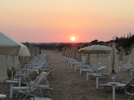}
\includegraphics[height=1.2cm]{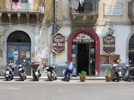}
\includegraphics[height=1.2cm]{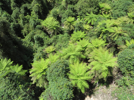}
\includegraphics[height=1.2cm]{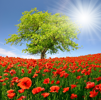}
\includegraphics[height=1.2cm]{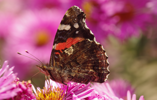}
\includegraphics[height=1.2cm]{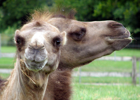}
\includegraphics[height=1.1cm]{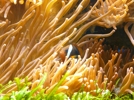}
\includegraphics[height=1.1cm]{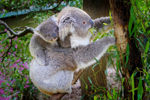}
\includegraphics[height=1.1cm]{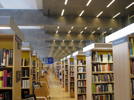}
\includegraphics[height=1.1cm]{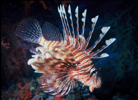}
\includegraphics[height=1.1cm]{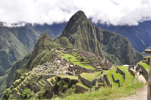}
\includegraphics[height=1.1cm]{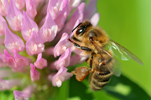}
\includegraphics[height=1.1cm]{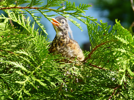}
\includegraphics[height=1.1cm]{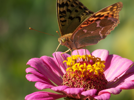}
\includegraphics[height=1.1cm]{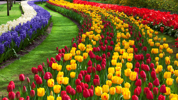}
\includegraphics[height=1.1cm]{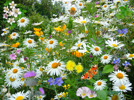}
}
\caption{L20 dataset. 20 high resolution large images.}
\label{fig:L20}
\end{figure*}

\noindent\textbf{L20 } is our newly proposed dataset. Since all the above mentioned datasets have images of medium-low resolution, below 0.5m pixels, we decided to created a new dataset, L20, with 20 large high resolution images. The images, as seen in Fig.~\ref{fig:L20}, are diverse in content, and their sizes vary from 3m pixels to up to 29m pixels. We conduct the self-similarity (S) experiments on the L20 dataset as discussed in Section~\ref{ssc:selfsimilarities}.

\subsection{Methods}
\label{ssc:methods}
We report results for a number of representative SR methods.

\noindent\textbf{Yang } is a method of Yang~\etal~\cite{Yang-CVPR-2008} that employs sparse coding and sparse dictionaries for learning a compact representation of the LR-HR priors/training samples and for sharp HR reconstruction results.

\noindent\textbf{Zeyde } is a method of Zeyde~\etal~\cite{Zeyde-CS-2012} that improves the Yang method by efficiently learning dictionaries using K-SVD~\cite{Aharon-TSP-2006} and employing Orthogonal Matching Pursuit (OMP) for sparse solutions.

\noindent\textbf{ANR } or Anchored Neighborhood Regression of Timofte~\etal~\cite{Timofte-ICCV-2013} relaxes the sparse decomposition optimization of patches from Yang and Zeyde to a ridge regression which can be solved offline and stored per each dictionary atom/anchor. This results in large speed benefits.

\noindent\textbf{A+ } of Timofte~\etal~\cite{Timofte-ACCV-2014} learns the regressors from all the training patches in the local neighborhood of the anchoring point/dictionary atom, and not solely from the anchoring points/dictionary atoms as ANR does. A+ and ANR have the same run-time complexity. See more in Section~\ref{sec:Aplus}.

\noindent\textbf{SRCNN } is a method introduced by Dong~\etal~\cite{Dong-ECCV-2014}, and is based on Convolutional Neural Networks (CNN)~\cite{LeCun-IEEE-1998}. It directly learns to map patches from low to high resolution images.

\subsection{Anchored regression baseline (A+)}
\label{sec:Aplus}
Our main baseline is the efficient Adjusted Anchored Neighborhood Regression (A+) method~\cite{Timofte-ACCV-2014}. The choice is motivated by the low time complexity of A+ both at training and testing and the superior performance that it achieves.
A+ and our improved methods share the same features for LR and HR patches and the same dictionary training (K-SVD~\cite{Aharon-TSP-2006}) as the ANR~\cite{Timofte-ICCV-2013} and Zeyde~\cite{Zeyde-CS-2012} methods.
The LR features are vertical and horizontal gradient responses, PCA projected for $99\%$ energy preservation.
The reference LR patch size is fixed to $3\times3$ while the HR patch is $s^2$ larger, with $s$ being the scaling factor, as in A+.

A+ assumes a partition of the LR space around the dictionary atoms, called anchors.
For each anchor $j$ a ridge regressor is trained on the local neighborhood ${\bf{N}}_l$ of fixed size of LR training patches (features). Thus, for each LR input patch ${\bf{y}}$ we minimize
\begin{equation}
\label{eqn:anr1}
\min_{\boldsymbol{\beta}}\lVert {\bf{y}} - {\bf{N}}_l {\boldsymbol{\beta}}\rVert ^2_2 + \lambda \lVert \boldsymbol{\beta} \rVert _2.
\end{equation}
Then the LR input patch ${\bf{y}}$ can be projected to the HR space as 
\begin{equation}
\label{eqn:anr2}
{\bf x} = {\bf{N}}_h ({\bf{N}}_l^T {\bf{N}}_l + \lambda {\bf{I}})^{-1} {\bf{N}}_l^T{\bf{y}} = {\bf{P}}_j {\bf{y}},
\end{equation}
where ${\bf{N}}_h$ are the corresponding HR patches of the LR neighborhood ${\bf{N}}_l$.  ${\bf{P}}_j$ is the stored projection matrix for the anchor $j$.
The SR process for A+ (and ANR) at test time then becomes a nearest anchor search followed by a matrix multiplication (application of the corresponding stored regressor) for each input LR patch.

\section{Proposed methods}
\label{sec:methods}

\subsection{Augmentation of training data (A)}
\label{ssc:augmentation}
 More training data results in an increase in performance up to a point where exponentially more data is necessary for any further improvement. This has been shown, among others, by Timofte~\etal in \cite{Timofte-ICCV-2013,Timofte-ACCV-2014} for neighbor embedding methods and anchored regression methods and by Dong~\etal in~\cite{Dong-ECCV-2014, Dong-PAMI-2015} for the convolutional neural networks-based methods.
 Zhu~\etal~\cite{Zhu-CVPR-2014} assume deformable patches and Huang~\etal~\cite{Huang-CVPR-2015} transform self-exemplars.

Around $\sim\!0.5$~million corresponding patches of $3\times3$ pixels for LR and $9\times9$ for HR are extracted from the Train91 images. By scaling the training images in~\cite{Timofte-ACCV-2014} 5 million patches are extracted for A+ and improve the PSNR performance from 32.39dB with 0.5 million to 32.55dB on Set5 and magnification $\times3$.
Inspired by the image classification literature~\cite{Chatfield-BMVC-2014}, we consider also the flipped and rotated versions of the training images/patches.
If we rotate the original images by $90^{\circ}$, $180^{\circ}$, $270^{\circ}$ and flip them upside-down (see Fig.~\ref{fig:augmentation}), we get 728 images without altered content.

\begin{figure}[]
    \centering
      {
        \begin{tikzpicture}
        \begin{axis}[
            width = \columnwidth,
    	    xmode = log,
            xmin=1000, xmax=100000000,
            ymin = 31.25, ymax = 33.25,
            xlabel= training samples,
            ylabel= PSNR (dB),
            legend cell align=right,
            legend pos= north west,
            ]
        
            \addplot[mark=*,magenta, ultra thick] plot coordinates {
            (    5000, 31.83)
            (   50000, 32.3)
    	    (  500000, 32.61)
    	    ( 5000000, 32.81)
    	    (50000000, 32.92)
            };
            \addplot[mark=square*,cyan, very thick, densely dashed] plot coordinates {
            (    5000, 31.82)
            (   50000, 32.28)
    	    (  500000, 32.57)
    	    ( 5000000, 32.76)
    	    (50000000, 32.85)
            };
            \addplot[mark=diamond*,green, thick, dashdotted] plot coordinates {
            (    5000, 31.79)
            (   50000, 32.25)
    	    (  500000, 32.43)
    	    ( 5000000, 32.59)
    	    (50000000, 32.71)
            };
            \addplot[mark=triangle*,red, thick, loosely dashed] plot coordinates {
    	    (    5000, 31.79)
            (   50000, 32.16)
    	    (  500000, 32.33)
    	    ( 5000000, 32.45)
    	    (50000000, 32.5)
            };
            \addplot[mark=star,blue, thick, dotted] plot coordinates {
            (    5000, 31.66)
            (   50000, 31.96)
    	    (  500000, 32.08)
    	    ( 5000000, 32.13)
    	    (50000000, 32.17)
            };
    
            \legend{{\bf 65536 regressors}\\8192 regressors\\1024 regressors\\128 regressors\\16 regressors\\}
    
            \end{axis}
        \end{tikzpicture}
    }
   \caption{Average PSNR performance of A+ on (Set5, $\times3$) improves with the number of training samples and regressors} 
    \label{fig:PSNR_vs_training}
\end{figure}
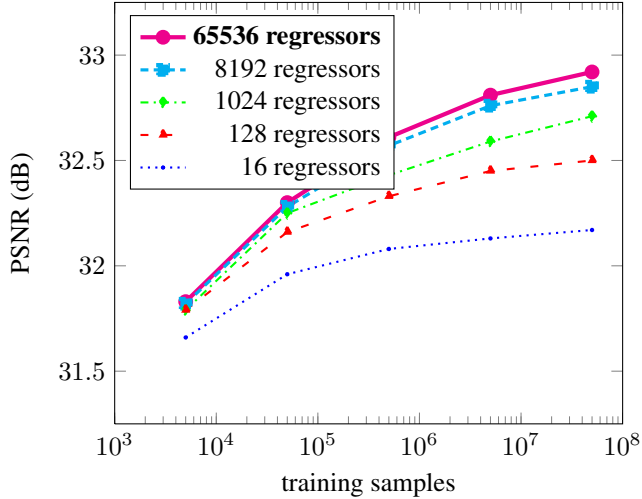

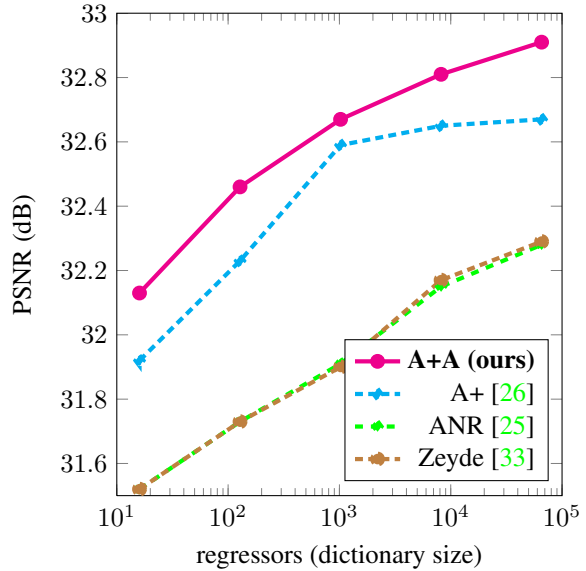
\begin{figure}[t!]
\centering
\begin{tabular}{c}
\begin{tikzpicture}
    \begin{axis}[
        width=0.9\columnwidth,
        height=8cm,
	    xmode = log, 
        xmin=10, xmax=100000,
        ymin = 31.5, ymax = 33.0,
        xlabel= regressors (dictionary size),
        ylabel= PSNR (dB),
        legend cell align=right,
        legend pos= south east,
    ]
    
        \addplot[mark=*,magenta, ultra thick] plot coordinates {
        (16, 32.13)
	(128, 32.46)
	(1024, 32.67)
        (8192, 32.81)
        (65536,32.91)    
        };
        \addplot[mark=diamond*,cyan, ultra thick, densely dashed] plot coordinates {
        (16, 31.92)
	    (128, 32.23)
	    (1024, 32.59)
        (8192, 32.65)
        (65536,32.67)    
        };

        \addplot[mark=diamond*,green, ultra thick, dashed] plot coordinates {
        (16, 31.52)
	    (128, 31.73)
	    (1024, 31.91)
        (8192, 32.15)
        (65536,32.28)    
        };

        \addplot[mark=*,brown, ultra thick, densely dashed] plot coordinates {
        (16, 31.52)
	    (128, 31.73)
	    (1024, 31.90)
        (8192, 32.17)
        (65536,32.29)    
        };

    \legend{{\bf A+A (ours)}\\A+~\cite{Timofte-ACCV-2014}\\ANR~\cite{Timofte-ICCV-2013}\\Zeyde~\cite{Zeyde-CS-2012}\\}
    \end{axis}
\end{tikzpicture}
\end{tabular}
\caption{Performance (Set5, $\times3$) improves with the number of regressors/atoms (dictionary size).}
\label{fig:PSNR_vs_atoms}
\end{figure}

In Fig.~\ref{fig:PSNR_vs_training} we show how the number of training LR-HR samples affects the PSNR performance of the A+ method on the Set5 images.
The performance of A+ with 1024 regressors varies from 31.83dB when trained on 5000 samples to 32.39dB for 0.5 million and 32.71dB when using 50 million training samples.
Note that the running time at test stays the same as it does not depend on the number of training samples but on the number of regressors. 
By \textbf{A+A} we mark our setup with A+, 65,536 regressors and 50 million training samples, improving 0.3dB over A+.

\subsection{Large dictionary and hierarchical search (H)}
\label{ssc:large}
In general, if the dictionary size (basis of samples/anchoring points) is increased, the performance for sparse coding (such as Zeyde~\cite{Zeyde-CS-2012} or Yang~\cite{Yang-CVPR-2008}) or anchored methods (such as ANR~\cite{Timofte-ICCV-2013} or A+~\cite{Timofte-ACCV-2014}) improves as the learned model generalizes better, as shown in Fig.~\ref{fig:PSNR_vs_atoms}.
We show in Fig.~\ref{fig:PSNR_vs_training} on Set5, $\times3$ how the performance of A+ increases when using 16 up to 65,536 regressors for any fixed size pool of training samples. In A+ each regressor is associated with an anchoring point. The anchors quantize the LR feature space. The more anchors are used, the smaller the quantization error gets and the easier is the local regression. On Set5 the PSNR is 32.17dB for 16 regressors, while it reaches 32.92dB for 65536 regressors with 50 million training samples (our \textbf{A+A} setup). However, the more regressors (anchors) are used, the slower the method gets. At running time, each LR patch (feature) is linearly matched to all the anchors. The regressor of the closest anchor is applied to reconstruct the HR patch. Obviously, this linear search in $\mathcal{O}(N)$ can be improved.
Yet, the LR features are high dimensional (30 after PCA reduction for $\times3$ for A+) and the speedup achievable with data search structures such as kd-trees, forests, or spherical hashing codes are rather small (3-4 times in~\cite{Perez-ACCV-2014,Schulter-CVPR-2015}).

Instead, we propose a hierarchical search structure in $\mathcal{O}(\sqrt{N})$ with very good empirical precision, that does not change the training procedure of A+. Given $N$ anchors and their $N$ regressors, we cluster them into $\sqrt{N}$ groups using k-means, each with an $l_2$-normalized centroid. To each centroid we assign the $c\sqrt{N}$ most correlated anchors. This results in a 2-layers structure. For each query we linearly search for the most correlated centroid ($1^{st}$ layer) to then linearly search within the anchors assigned to it ($2^{nd}$ layer). $c=4$ is fixed in all our experiments, so that one anchor potentially can be assigned to more centroids to handle the cluster boundary well.

\begin{figure}[t!]
\centering
\begin{tabular}{c}
\begin{tikzpicture}
    \begin{axis}[
        width=\columnwidth,
        height=6cm,
        xmode = log, 
        xmin=0, xmax=10,
        ymin = 32, ymax = 33.0,
        xlabel= encoding time (s),
        ylabel= PSNR (dB),
        legend cell align=right,
        legend pos= south east,
        scatter/classes={%
		    1={mark=*,magenta},%
		    2={mark=diamond*,cyan}}        
    ]
    
        \addplot[mark=*,magenta, ultra thick] plot coordinates {
        (0.05, 32.13)
	    (0.13, 32.46)
	    (0.21, 32.67)
        (0.59, 32.80)
        (2.98, 32.91)    
        };
        \addplot[mark=diamond*,cyan, ultra thick, densely dashed] plot coordinates {
        (0.046, 32.12)
	    (0.1 , 32.43)
	    (0.15, 32.64)
        (0.20, 32.78)
        (0.22, 32.89)    
        };
       
       	\addplot[scatter, mark=*,mark size=0.1,mark options={fill=white}, only marks,nodes near coords, point meta=explicit symbolic, visualization depends on={value \thisrow{anchor}\as\myanchor},           every node near coord/.append style={font = \footnotesize, inner sep = 2pt,  color = magenta, anchor=\myanchor}, forget plot]  table [x= y, y= x, meta=myvalue] {psnr_regressors.txt};
		
		\addplot[scatter, mark=*,mark size=0.1, mark options={fill=white}, only marks,nodes near coords,
           point meta=explicit symbolic,
           visualization depends on={value \thisrow{anchor}\as\myanchor},
           every node near coord/.append style={font = \footnotesize, inner sep = 3pt, color = cyan, anchor=\myanchor}, forget plot]      
		    table [x= y, y= x, meta=myvalue1] {psnr_regressors.txt};       
      \legend{linear search structure\\hierarchical search structure\\}
    \end{axis}
\end{tikzpicture}
\end{tabular}
\caption{Performance (A+A on Set5, $\times3$) depends on the number of regressors and the data search structure.}
\label{fig:PSNR_vs_regressors}
\vspace{-0.5cm}
\end{figure}
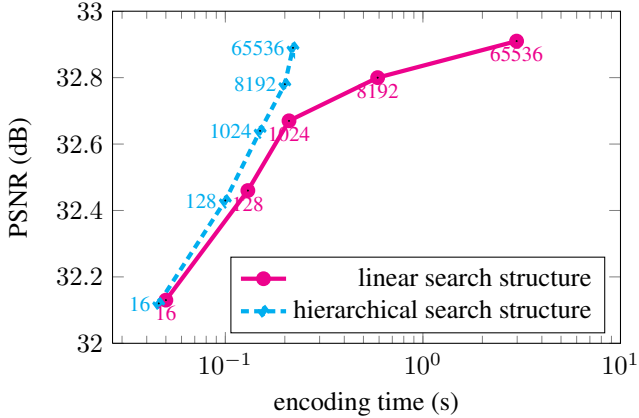

In Fig.~\ref{fig:PSNR_vs_regressors} we depict the performance of A+A with and without our hierarchical search structure in relation to the number of trained regressors. The hierarchical structure looses at most only 0.03dB but consistently speeds up above 1,024 regressors. A+A with hierarchical search (H) and 65,536 regressors has a running time comparable to the original A+ with linear search and 1,024 regressors, but is 0.3dB better. 

\subsection{Back projection (B)}
\label{ssc:backprojection}
Applying an iterative back projection (B) refinement~\cite{Irani-CVGIP-1991} generally improves the PSNR as it makes the HR reconstruction consistent with the LR input and the employed degradation operators such as blur, downscaling, and downsampling.
Knowing the degradation operators is a must for the IBP approaches and therefore they need to be estimated~\cite{Michaeli-ICCV-2013}.
Assuming the degradation operators to be known, (B) improves the PSNR of A+ by up to 0.06dB, depending on the settings as shown in column \textbf{A+B} in Table~\ref{tab:avg_results}, when starting from the A+ results. 

In Table~\ref{tab:PSNR_vs_IBP} we compare the improvements obtained with our iterative back projection (B) refinement when starting from different SR methods. The largest improvement is of 0.59dB when starting from the sparse coding method of Yang~\etal~\cite{Yang-CVPR-2008}, whereas for A+ it only improves 0.04dB. This behaviour can be explained by the fact that the reference A+ is 1.18dB better than the reference Yang method. Therefore A+'s HR reconstruction is much more consistent with the LR image than Yang's and improving by using (B) is more difficult. The refined Yang result is 0.59dB better than the baseline Yang method but still 0.59dB behind A+ without refinement.
Note that generally the degradation operators are unknown and their estimation is not precise, therefore our reported results with (B) refinement are an upper bound for a practical implementation and difficult to reach.

\begin{table}[]
\caption{Back projection (B) improves the super-resolution PSNR results (Set5, $\times3$).}
\centering
\setlength{\tabcolsep}{3pt}
\begin{tabular}{c||ccccc|c }
 {\bf (B)}  &  Yang & ANR & Zeyde & SRCNN & A+ & IA\\
 & \cite{Yang-CVPR-2008}& \cite{Timofte-ICCV-2013} & \cite{Zeyde-CS-2012} & \cite{Dong-ECCV-2014}&\cite{Timofte-ACCV-2014}&(ours)\\
  \hline
  \hline
 $\times$  & 31.41 & 31.92 & 31.90 & 32.39 & 32.59 & 33.46\\
 \checkmark & 32.00 & 31.99 & 32.04 & 32.52 & 32.63 & 33.51\\
  \hline
  \hline
  \textcolor{red}{\bf Improv.} & \textcolor{red}{\bf +0.59} & \textcolor{red}{\bf +0.07} &\textcolor{red}{\bf +0.14} & \textcolor{red}{\bf +0.13} & \textcolor{red}{\bf +0.04} & \textcolor{red}{\bf +0.05}\\
\end{tabular}
\label{tab:PSNR_vs_IBP}
\vspace{-0.5cm}
\end{table}

\subsection{Cascade of anchored regressors (C)}
\label{ssc:cascade}
As the magnification factor is decreased, the super-resolution becomes more accurate, since the space of possible HR solutions for each LR patch and thus the ambiguity decreases.
Glasner~\etal~\cite{Glasner-ICCV-2009} use small SR steps to gradually refine the contents up to the desired HR. The errors are usually enlarged by the subsequent steps and the time complexity depends on the number of steps.
Instead of super-resolving the LR image in small steps, we can go in one step (stage) and then refine the prediction using the same SR method again adapted to this input. We consider the output of the previous stage as LR image input and as target the HR image for each stage.
Thus, we build a cascade of trained models, where each stage brings the prediction closer to the target HR image. The cascades and the layered or recurrent processing are broadly used concepts in vision tasks (\ie object detection~\cite{Viola-CVPR-2001} and deep learning~\cite{Chatfield-BMVC-2014}). The method of Peleg and Elad~\cite{Peleg-TIP-2014} and the SRCNN method of Dong~\etal~\cite{Dong-ECCV-2014} are layered by design.
While the incremental approach has a loose control over the errors, the cascade explicitly minimizes the prediction errors at each stage.

\begin{figure}[t!]
    \centering
        {
        \begin{tikzpicture}
            \begin{axis}[
            width = \columnwidth,
            xmin  = 0.5, xmax=5.5,
            ymin  = 32.9, ymax = 33.5,
            xlabel= stages in cascade,
            ylabel= PSNR (dB),
            legend cell align=right,
            legend pos= south east,
            ]
        
            \addplot[mark=*,green, ultra thick] plot coordinates {
            (1, 32.97)
        	(2, 33.24)
        	(3, 33.36)
        	(4, 33.44)
            (5, 33.45)        
            };
            \addplot[mark=square*,blue, ultra thick, dashed] plot coordinates {
            (1, 32.92)
        	(2, 33.14)
        	(3, 33.19)
        	(4, 33.21)
            (5, 33.21)        
            };
    
    \legend{with enhanced prediction\\without enhanced prediction\\}
            \end{axis}
        \end{tikzpicture}
        }
   \caption{Average PSNR performance of IA improves with the number of cascade stages (Set5, $\times3$).} 
    \label{fig:PSNR_vs_cascade}
    \vspace{-0.25cm}
\end{figure}
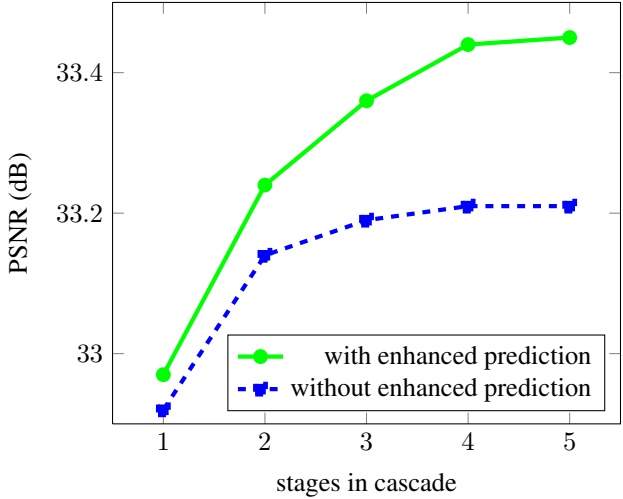

In our cascaded A+, called \textbf{A+C}, with 50 million training samples, we keep the same features and settings for all the stages but have models that have been trained per stage.
As shown in Fig.~\ref{fig:PSNR_vs_cascade} and Table~\ref{tab:PSNR_vs_C} the performance improves from 32.92dB after the $1^{st}$ stage of the cascade and saturates at 33.21dB after the $4^{th}$ stage of the cascade. The running time is linear in the number of stages.
The same cascading idea of A+ was applied for image demosaicing in~\cite{Wu-ICIP-2015} with two stages.

\begin{table}[t!]
\caption{Cascading (C) and enhanced prediction (E) improve the super-resolution PSNR results (Set5, $\times3$).}
\centering
\begin{tabular}{l|c||ccc|c }
 {\bf Cascade}  & {\bf (E)} &  ANR & Zeyde & A+ & IA\\
 && \cite{Timofte-ICCV-2013} & \cite{Zeyde-CS-2012} & \cite{Timofte-ACCV-2014}&(ours)\\
 \hline
 \hline
 1 stage  &  $\times$ & 31.92 & 31.90 & 32.59 & 32.77\\
 1 stage  & \checkmark& 32.15 & 32.16 & 32.70 & 32.91\\
  \hline
 2 stages &  $\times$ & 32.19 & 32.20 & 32.70 & 33.05\\
 2 stages & \checkmark& 32.25 & 32.26 & 32.81 & 33.21\\
 \hline
 3 stages &  $\times$ & 32.22 & 32.23 & 32.76 & 33.18\\
 3 stages & \checkmark& 32.28 & 32.29 & 32.87 & 33.34\\
 \hline
 4 stages &  $\times$ & 32.24 & 32.24 & 32.79 & 33.33\\
 4 stages & \checkmark& 32.30 & 32.31 & 32.89 & 33.46\\
 \hline
 \hline
  \multicolumn{2}{c||}{\textcolor{red}{\bf Improvement}} &\textcolor{red}{\bf +0.38} & \textcolor{red}{\bf +0.41} & \textcolor{red}{\bf +0.30} & \textcolor{red}{\bf +0.69}\\
\end{tabular}
\label{tab:PSNR_vs_C}
\end{table}

\begin{table}[]
\caption{Enhanced prediction (E) improves the super-resolution PSNR results (Set5, $\times3$).}
\centering
\setlength{\tabcolsep}{3pt}
\begin{tabular}{c||ccccc|c }
{\bf (E)}  &  Yang & ANR & Zeyde & SRCNN & A+ & IA\\
 & \cite{Yang-CVPR-2008}& \cite{Timofte-ICCV-2013} & \cite{Zeyde-CS-2012} & \cite{Dong-ECCV-2014}&\cite{Timofte-ACCV-2014}&(ours)\\
 \hline
 \hline
 $\times$  & 31.41 & 31.92 & 31.90 & 32.39 & 32.59 & 33.21\\
 \checkmark & 31.65 & 31.97 & 31.96 & 32.61 & 32.68 & 33.46\\
  \hline
    \hline
  \textcolor{red}{\bf Improv.} & \textcolor{red}{\bf +0.24} & \textcolor{red}{\bf +0.05} &\textcolor{red}{\bf +0.06} & \textcolor{red}{\bf +0.22} & \textcolor{red}{\bf +0.09} & \textcolor{red}{\bf +0.25}\\
\end{tabular}
\label{tab:PSNR_vs_E}
\vspace{-0.5cm}
\end{table}

\subsection{Enhanced prediction (E)}
\label{ssc:enhancedprediction}
In image classification~\cite{Chatfield-BMVC-2014} often the prediction for an input image is enhanced by averaging the predictions on a set of transformed images derived from it.
The most common transformations include cropping, flipping, and rotations.
In SR image rotations and flips should lead to the same HR results at pixel level. Therefore, we apply rotations and flips on the LR image as shown in see Fig.~\ref{fig:augmentation} to get a set of 8 LR images, then apply the SR method on each, reverse the transformation on the HR outputs and average for the final HR result.

On Set5 (see Fig.~\ref{fig:PSNR_vs_cascade} and Table~\ref{tab:PSNR_vs_C}) the enhanced prediction (E) gives a 0.05dB improvement for a single stage and more than 0.24dB when 4 stages are employed in the cascade.
The running time is linear in the number of transformations.
In Table~\ref{tab:PSNR_vs_E} we report the improvements due to (E) for different SR methods.
It varies from +0.05dB for ANR up to +0.25dB for the Yang method.

\begin{figure}[t!]
    \centering
  {
        \begin{tikzpicture}
        \begin{axis}[
        width = \columnwidth,
    	xmode = log, 
            xmin  = 10000, xmax=10000000,
            ymin  = -0.2, ymax = 0.1,
            xlabel= image size (pixels),
            ylabel= PSNR (dB) gain,
            legend cell align=left,
            legend pos= south east,
        ]
        
            \addplot[mark=*,red, ultra thick, solid] plot coordinates {
            (15075,  26.435-26.427)
            (60299,  28.173-28.146)
            (241198, 30.062-30.00)
            (964791, 32.359-32.29)
            (3859164,35.832-35.75)
            };
            \addplot[mark=diamond*,blue, ultra thick, dashed] plot coordinates {
            (15075,  25.974-26.427)
            (60299,  28.03-28.146)
            (241198, 30.013-30.00)
            (964791, 32.32-32.29)
            (3859164, 35.81-35.75)
            };
            \addplot[mark=square*,green, ultra thick, loosely dotted] plot coordinates {
            (15075,  0)
            (60299,  0)
            (241198, 0)
            (964791, 0)
            (3859164,0)
            };
            \legend{combined\\internal dictionary\\external dictionary (reference)\\}
    
        \end{axis}
        \end{tikzpicture}
  }
   \caption{Average PSNR gain comparison of internal dictionary, external dictionary and combined dictionary with respect to the input LR image size (L20, $\times3$).} 
    \label{fig:PSNR_vs_dictionary}
    \vspace{-0.5cm}
\end{figure}
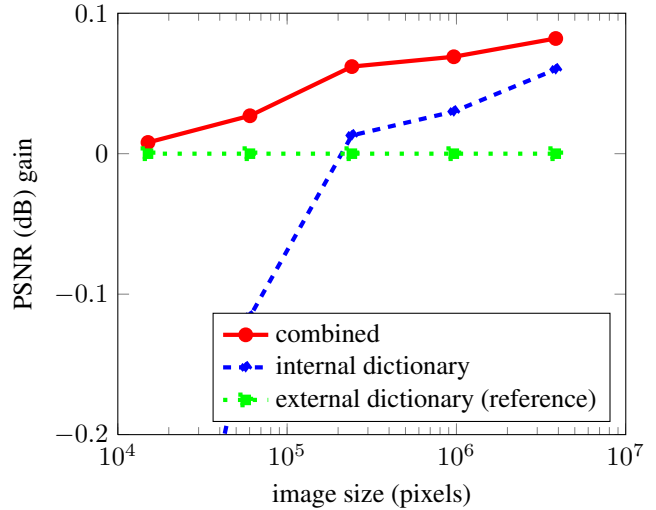

\subsection{Self-similarities (S)}
\label{ssc:selfsimilarities}
The image self-similarities (or patch redundancy) at different image scales can help to discriminate between equally possible HR reconstructions.
While we considered external dictionaries, thus priors from LR and HR training images,
some advocate internal dictionaries, \ie dictionaries built from the input LR image, matching the image context. Exponents are Glasner~\etal~\cite{Glasner-ICCV-2009} or Dong~\etal~\cite{Dong-TIP-2013}, among others~\cite{Yang-ACCV-2010-short,Freedman-TG-2011,Yang-CVPR-2013,Huang-CVPR-2015}. Extracting and building models adapted to each new input image is expensive. Also, in recent works, and on the standard benchmarks, methods such as SRCNN~\cite{Dong-ECCV-2014} and A+~\cite{Timofte-ACCV-2014} based on external dictionaries proved better in terms of PSNR and running time.

We point out that depending on the size of the input LR image and the textural complexity, the internal dictionaries can be better than the external dictionaries. Huang~\etal~\cite{Huang-CVPR-2015} report better results with internal dictionaries when using urban HR images with high geometric regularities.
We downsize the L20 images and plot the improvements over an external dictionary in Fig.~\ref{fig:PSNR_vs_dictionary}. Above $246,000$ LR pixels the internal dictionary improves over the external one. However, the best results are obtained using both external and internal dictionaries.

\subsection{Reasoning with context (R)}
\label{ssc:context}
Intuition says that using the immediate surrounding of a LR patch should help. For example, Dong~\etal~\cite{Dong-TIP-2011} train domain specific models and Sun~\etal~\cite{Sun-CVPR-2010} hallucinate using context constraints.
We consider context images centered on each LR patch of size equal with the LR patch size times the scaling factor ($\times3$).
We extract the same features as for the LR patches but in the ($\times3$) downscaled image and cluster them into 4 groups with 4 centroids. A small number that does not increase the time complexity much but it is still relevant for analyzing the context idea. 
We keep the standard A+ pipeline with 1024 anchors and 0.5 million training patches ( A+(0.5m) ). To each anchor we assign the closest patches and instead of training one regressor as A+ would, we train 4 context specific regressors. For each context we compute a regressor using the 1024 patches closest to both anchor and context centroid, in a 10 to 1 contribution. For patches of comparable distances to the anchor the distance to the context centroid makes the difference.
At testing time, each LR patch is first matched against the anchors and then the regressor of the closest context is used to get the HR output.
By reasoning with context we improve from 32.39dB to 32.55dB on Set5, while the running time only slightly increases. 
In Table~\ref{tab:PSNR_vs_R} we report the improvements achieved using reasoning with context (R) over original SR methods. The (R) derivations were similar to the one explained for the A+ (0.5m) setup.

\begin{table}[t!]
\caption{Reasoning with context (R) improves the super-resolution PSNR results (Set5, $\times3$).}
\centering
\begin{tabular}{c|cccc|c }
 {\bf (R)}  &  ANR & Zeyde & A+{\footnotesize (0.5m)} & A+  & IA\\
 Context& \cite{Timofte-ICCV-2013} & \cite{Zeyde-CS-2012} & \cite{Timofte-ACCV-2014} &\cite{Timofte-ACCV-2014}&(ours)\\
 \hline
 \hline
  $\times$     & 31.92 & 31.90 & 32.39& 32.59 & 33.46\\
 \checkmark    & 32.12 & 32.11 & 32.55& 32.71 & 33.51\\
 \hline
 \hline
  \textcolor{red}{\bf Improv.} &\textcolor{red}{\bf +0.20} & \textcolor{red}{\bf +0.21} & \textcolor{red}{\bf +0.16} & \textcolor{red}{\bf +0.12} & \textcolor{red}{\bf +0.05}\\
\end{tabular}
\label{tab:PSNR_vs_R}
\vspace{-0.5cm}
\end{table}

\subsection{Improved A+ (IA)}
\label{ssc:IA}
Any combination of the proposed techniques would likely improve over the baseline example-based super-resolution method.
If we start from the A+ method, and (A) add augmentation (50 million training samples), increase the number of regressors (to 65536) and (H) use the hierarchical search structure, we achieve 0.33dB improvement over A+ (Set5, $\times3$) without an increase in running time.
Adding reasoning with context (R) slightly increases the running time for a gain of 0.1dB.
The cascade (C) allows for another jump in performance, +0.27dB, while the enhanced prediction (E) brings another 0.25dB. The gain brought by (C) and (E) comes at the price of increasing the computation time. The full setup, using (A, H, R, C, E) is marked as our proposed Improved A+ (IA) method. The addition of internal dictionaries (S) is possible but undesirable due to the computational cost.
Adding IBP (B) to the IA method can further improve the performance by 0.05dB.

\begin{figure}
\centering
\includegraphics[width=0.9\linewidth]{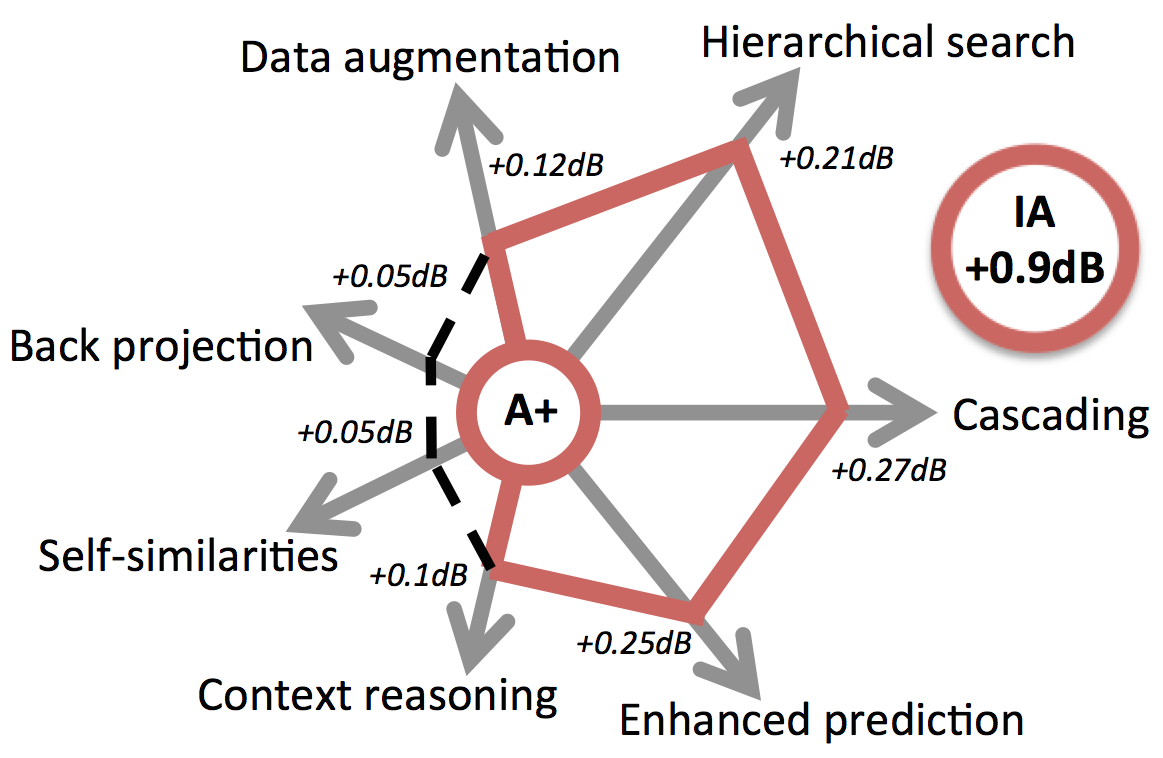}
\caption{Seven ways to Improve A+. PSNR gains for Set5, $\times3$.}
\label{fig:seven}
\vspace{-0.25cm}
\end{figure}

The seven ways to improve A+ are summarized in Fig.~\ref{fig:seven}. The Improved A+ (IA) method is 0.9dB better than the baseline A+ method by using 5 techniques (A, H, R, C, E).

Table~\ref{tab:avg_results} compares the results with A+~\cite{Timofte-ACCV-2014}, Zeyde~\cite{Zeyde-CS-2012}, and SRCNN~\cite{Dong-ECCV-2014} on standard benchmarks and for magnifications $\times2$, $\times3$, $\times4$. Figs.~\ref{fig:visual_crops} and~\ref{fig:visual_results} show visual results.

\begin{table*}
\vspace{-0.2cm}
 \setlength{\tabcolsep}{4pt} 
\centering
\caption{Average PSNR on Set5, Set14, and B100 and the \textcolor{red}{improvement (red)} of \textcolor{blue}{our IA (blue)} over \textbf{A+ (bold)} method.
}
\vspace{0.01cm}
{
\begin{tabular}{lc||c c c c c c|c c c c|c}
\multicolumn{2}{c||}{Benchmark}&Bicubic&NE+LLE&Zeyde&ANR& SRCNN&\textbf{A+}&A+B&A+A&A+C&\textcolor{blue}{\bf IA}&\textcolor{red}{\bf Improvement}\\
\multicolumn{2}{c||}{ } &&\cite{Timofte-ICCV-2013}&\cite{Zeyde-CS-2012}&\cite{Timofte-ICCV-2013}& \cite{Dong-ECCV-2014}&\cite{Timofte-ACCV-2014}&(ours)&(ours)&(ours) &\textcolor{blue}{\bf (ours)}&\textcolor{red}{\bf of IA over A+}\\
\hline\hline
              & x2 & 33.66 & 35.77 & 35.78 & 35.83 & 36.34 &\textbf{36.55}&36.60&36.89&37.26&\textcolor{blue}{\bf 37.39}&\textcolor{red}{\bf +0.84}\\
\textbf{Set5 }& x3 & 30.39 & 31.84 & 31.90 & 31.92 & 32.39 &\textbf{32.59}&32.63&32.92&33.20&\textcolor{blue}{\bf 33.46}&\textcolor{red}{\bf+0.87}\\
              & x4 & 28.42 & 29.61 & 29.69 & 29.69 & 30.09 &\textbf{30.29}&30.33&30.58&30.86&\textcolor{blue}{\bf 31.10}&\textcolor{red}{\bf+0.81}\\
\hline
              & x2 & 30.23 & 31.76 & 31.81 & 31.80 & 32.18 &\textbf{32.28}&32.33&32.48&32.73&\textcolor{blue}{\bf 32.87}&\textcolor{red}{\bf+0.59}\\
\textbf{Set14}& x3 & 27.54 & 28.60 & 28.67 & 28.65 & 29.00 &\textbf{29.13}&29.16&29.33&29.51&\textcolor{blue}{\bf 29.69}&\textcolor{red}{\bf+0.56}\\
              & x4 & 26.00 & 26.81 & 26.88 & 26.85 & 27.20 &\textbf{27.32}&27.35&27.54&27.74&\textcolor{blue}{\bf 27.88}&\textcolor{red}{\bf+0.56}\\
              
\hline

             & x2 & 29.32 & 30.41 & 30.40 & 30.44 & 30.71 &\textbf{30.78}&30.81&30.91&31.15&\textcolor{blue}{\bf 31.33}&\textcolor{red}{\bf+0.55}\\
\textbf{B100}& x3 & 27.15 & 27.85 & 27.87 & 27.89 & 28.10 &\textbf{28.18}&28.20&28.32&28.45&\textcolor{blue}{\bf 28.58}&\textcolor{red}{\bf+0.40}\\
             & x4 & 25.92 & 26.47 & 26.51 & 26.51 & 26.66 &\textbf{26.77}&26.79&26.91&27.03&\textcolor{blue}{\bf 27.16}&\textcolor{red}{\bf+0.39}\\

\end{tabular}
}
\label{tab:avg_results}
\vspace{-0.5cm}
\end{table*}

\section{Discussion}
\label{sec:discussion}

\subsection{Generality of the seven ways}
\label{ssc:generality}
Our study focused and demonstrated the seven ways to improve SR mainly on the A+ method.
As a result, the IA method has been proposed, combining 5 out of 7 ways, namely (A, H, R, C, E).
The effect of applying the different techniques is additive, each contributing to the final performance.
These techniques are general in the sense that they can be applied to other example-based single image super-resolution methods as well. We demonstrated the techniques on five methods.

In Fig.~\ref{fig:teaser} we report on a running time versus PSNR performance scale the results (Set5, $\times3$) of the reference methods A+, ANR, Zeyde, and Yang together with the improved results starting from these methods.
The A+A method combines A+ with A and H, while the A+C method combines A+ with A, H, and C. A+A and A+C are lighter versions of our IA.
For the improved ANR result we combined the A, H, R, B, and E techniques, for the improved Zeyde result we combined A, R, B, and E, while for Yang we combined B and E without retraining the original model.

Note that using combinations of the seven techniques we are able to improve significantly all the methods considered in our study which validates the wide applicability of these techniques. Thus, A+ is improved by 0.9dB in PSNR, Yang and ANR by 0.8dB and Zeyde by 0.7dB.

\subsection{Benchmark results}
\label{ssc:benchmarks}
All the experiments until now used Set5 and L20 and magnification factor $\times3$.
In Table~\ref{tab:avg_results} we report the average PSNR performance on Set5, Set14, and B100, and for magnification factors $\times2$, $\times3$, and $\times4$ of our methods in comparison with the baseline A+~\cite{Timofte-ACCV-2014}, ANR~\cite{Timofte-ICCV-2013}, Zeyde~\cite{Zeyde-CS-2012}, and SRCNN~\cite{Dong-ECCV-2014} methods. Also we report the result of the bicubic interpolation and the one for the Neighbor Embedding with Locally Linear Embedding (NE+LLE) method of Chang~\etal~\cite{Chang-CVPR-2004} as adapted and implemented in~\cite{Timofte-ICCV-2013}. All the methods used the same Train91 dataset for training.
For reporting improved results also for magnification factors $\times2$ and $\times4$, we keep the same parameters/settings as used for the case of magnification $\times3$ for our A+B, A+A, A+C, and IA methods.
A+B is provided for reference as the degradation operators usually are not known and difficult to estimate in practice. A+B just slightly improves over A+.
A+A improves 0.13dB up to 0.34dB over A+ while preserving the running time. A+C further improves at the price of running time, using a cascade with 3 stages.
IA improves 0.4dB up to 0.9dB over the A+ results, and significantly more over SRCNN, Zeyde, and ANR. 

\subsection{Qualitative assessment}
\begin{figure}
\centering
\setlength{\tabcolsep}{1pt}     
\renewcommand{\arraystretch}{0.2} 
\resizebox{\columnwidth}{!}{
\begin{tabular}{cccc}
Input & Zeyde~\cite{Zeyde-CS-2012} & A+~\cite{Timofte-ACCV-2014} & \textbf{IA (ours)}\\
\includegraphics[width=0.075\linewidth]{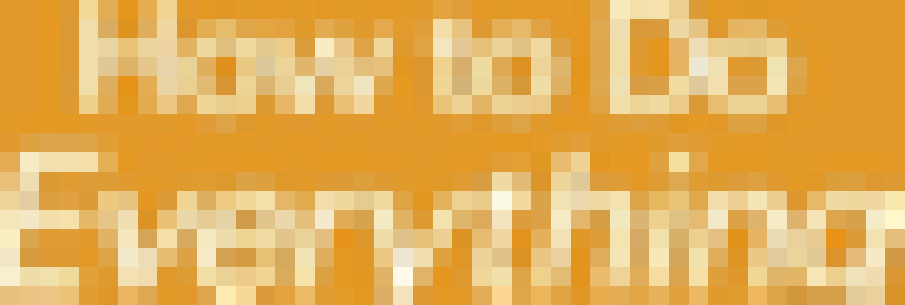}&
\includegraphics[width=0.3\linewidth]{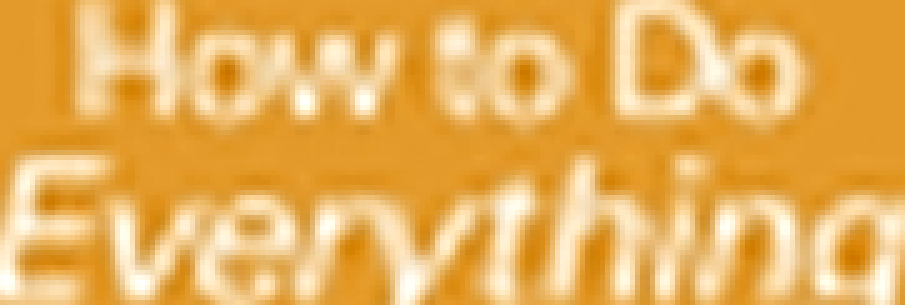}&
\includegraphics[width=0.3\linewidth]{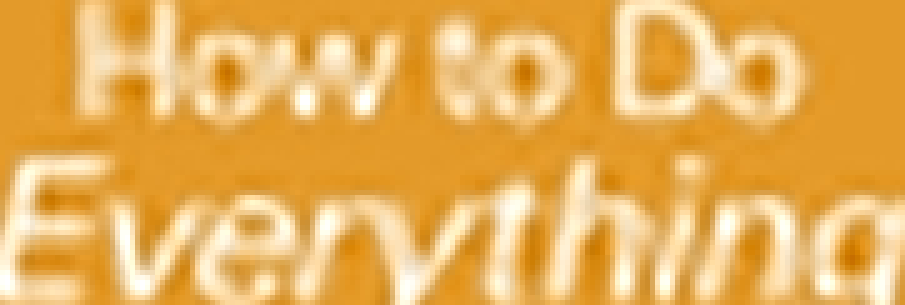}&
\includegraphics[width=0.3\linewidth]{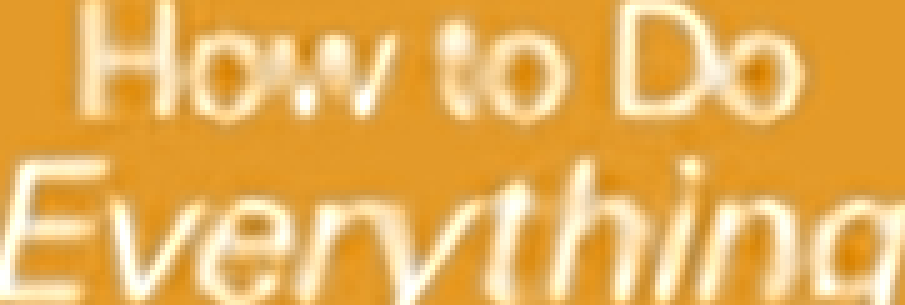}\\
\includegraphics[width=0.075\linewidth]{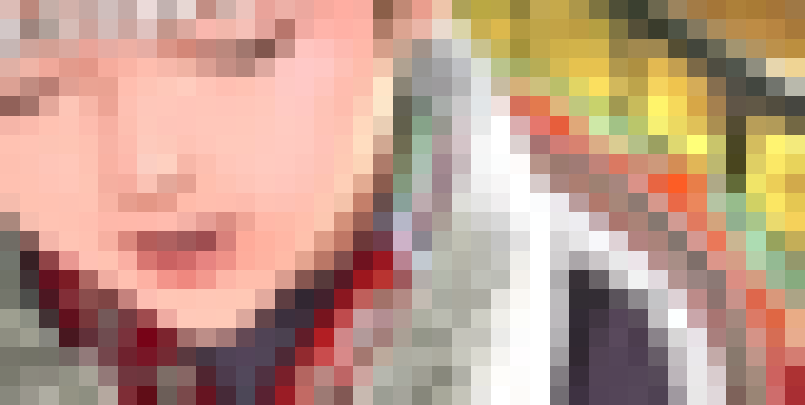}&
\includegraphics[width=0.3\linewidth]{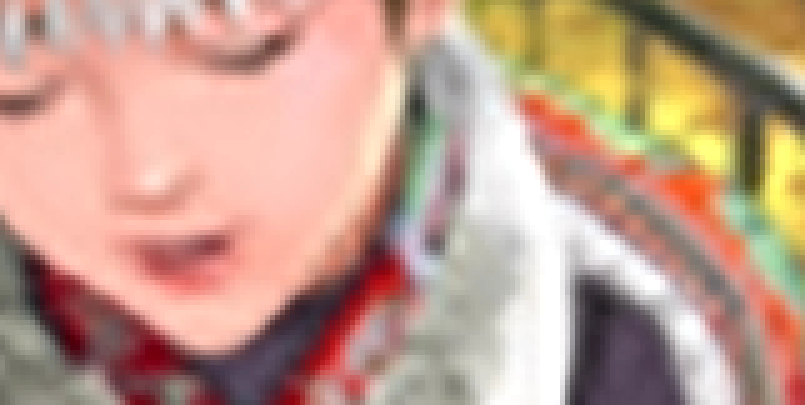}&
\includegraphics[width=0.3\linewidth]{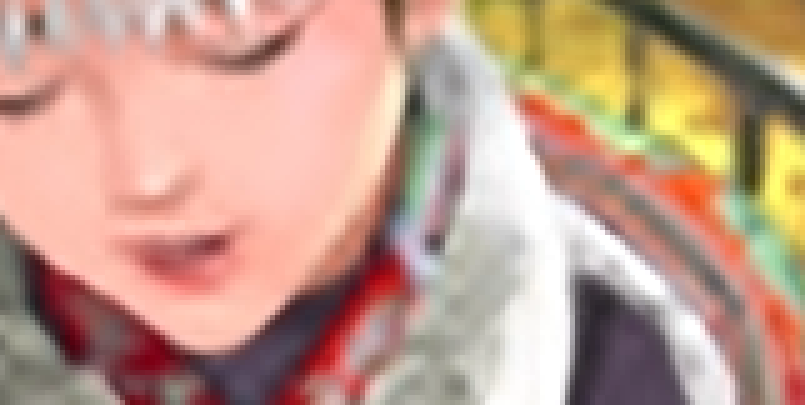}&
\includegraphics[width=0.3\linewidth]{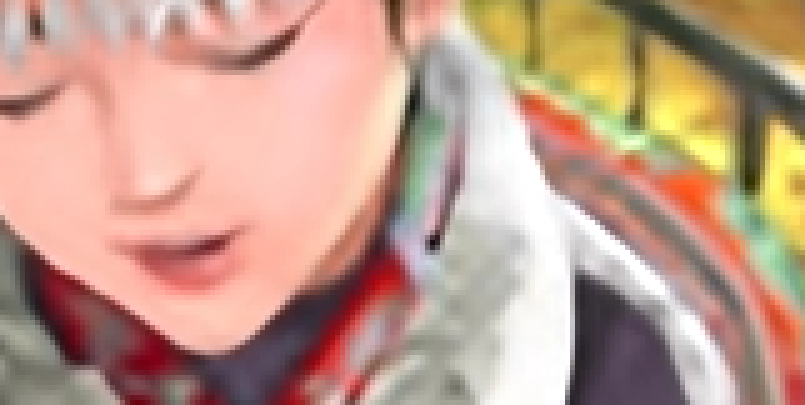}\\
\includegraphics[width=0.075\linewidth]{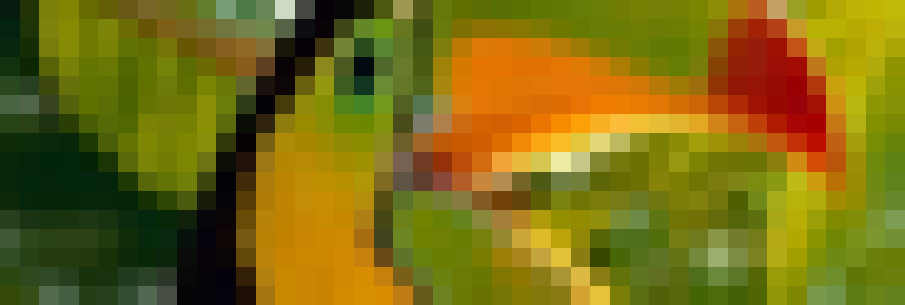}&
\includegraphics[width=0.3\linewidth]{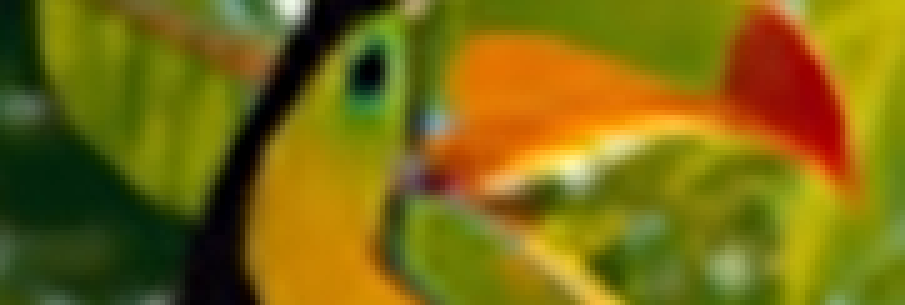}&
\includegraphics[width=0.3\linewidth]{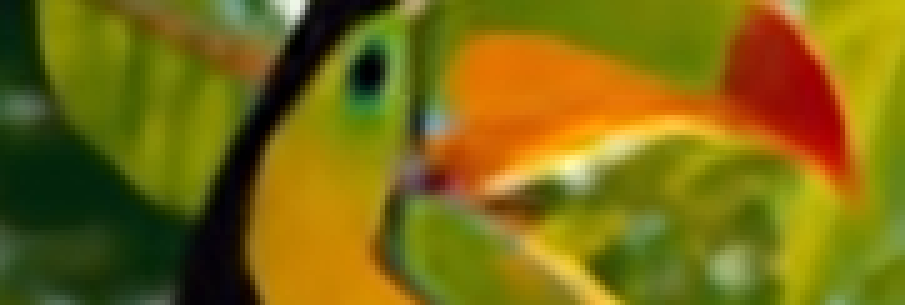}&
\includegraphics[width=0.3\linewidth]{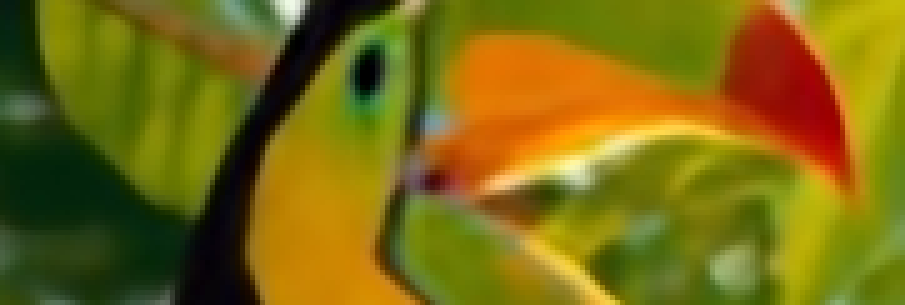}\\
\includegraphics[width=0.075\linewidth]{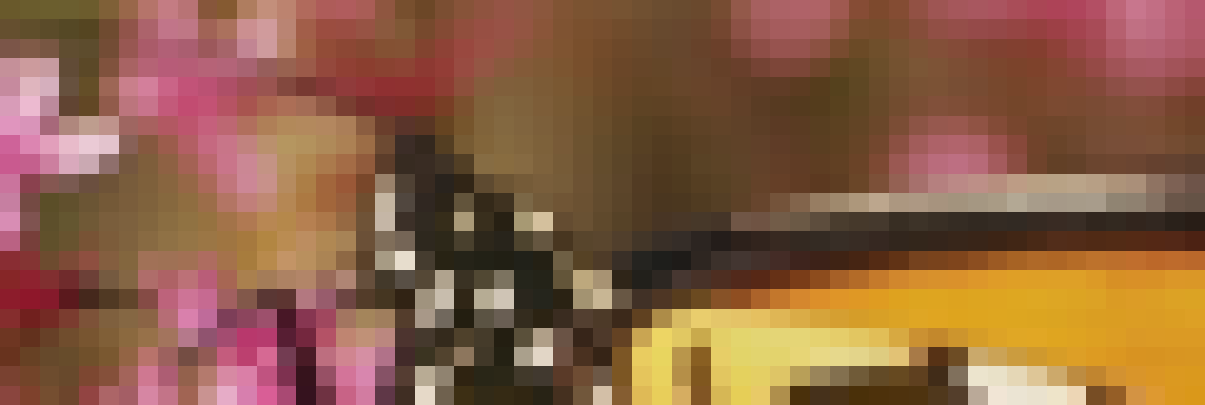}&
\includegraphics[width=0.3\linewidth]{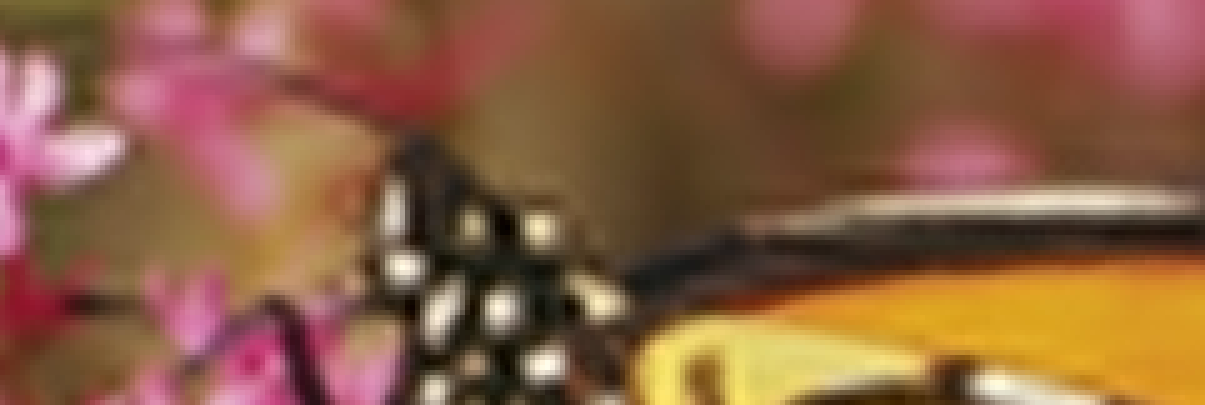}&
\includegraphics[width=0.3\linewidth]{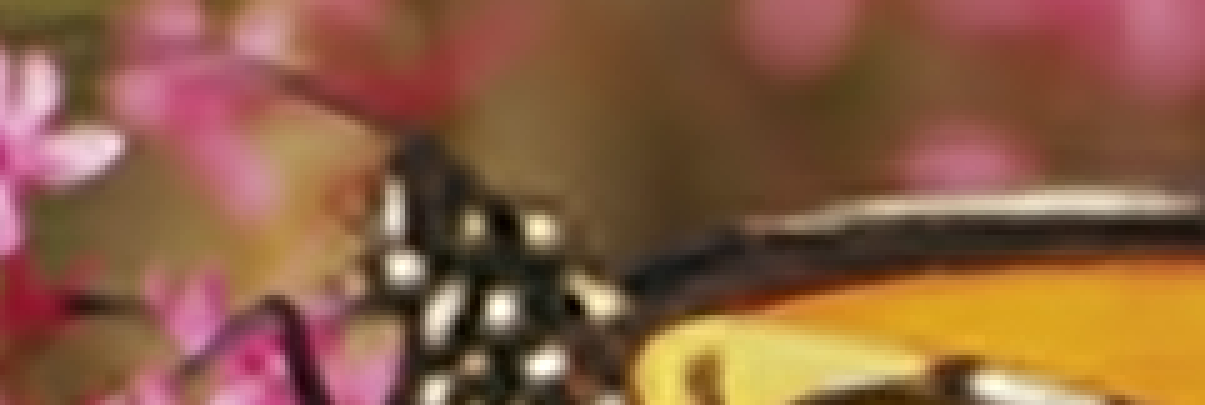}&
\includegraphics[width=0.3\linewidth]{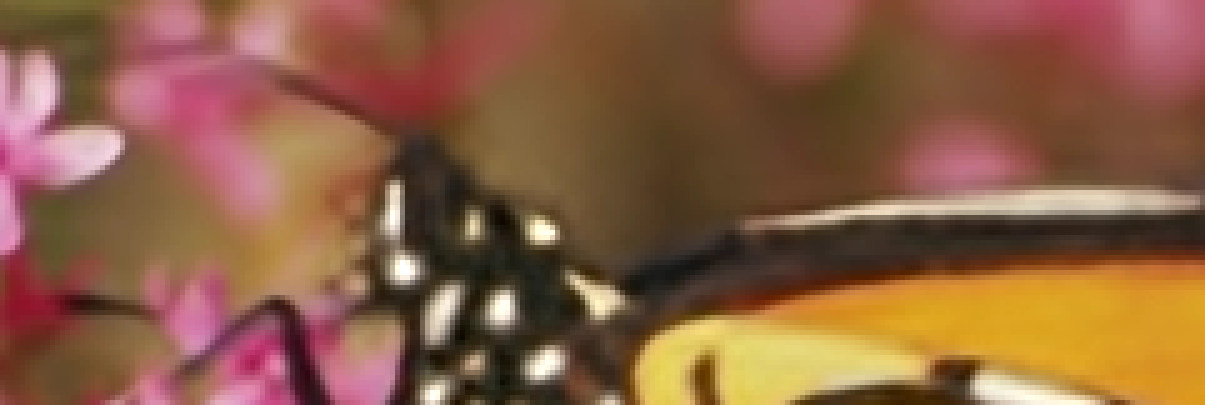}\\
\includegraphics[width=0.075\linewidth]{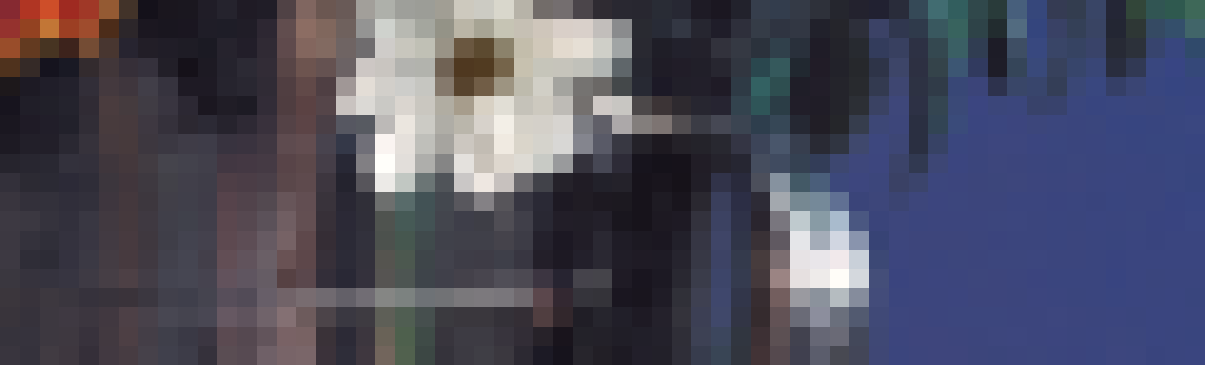}&
\includegraphics[width=0.3\linewidth]{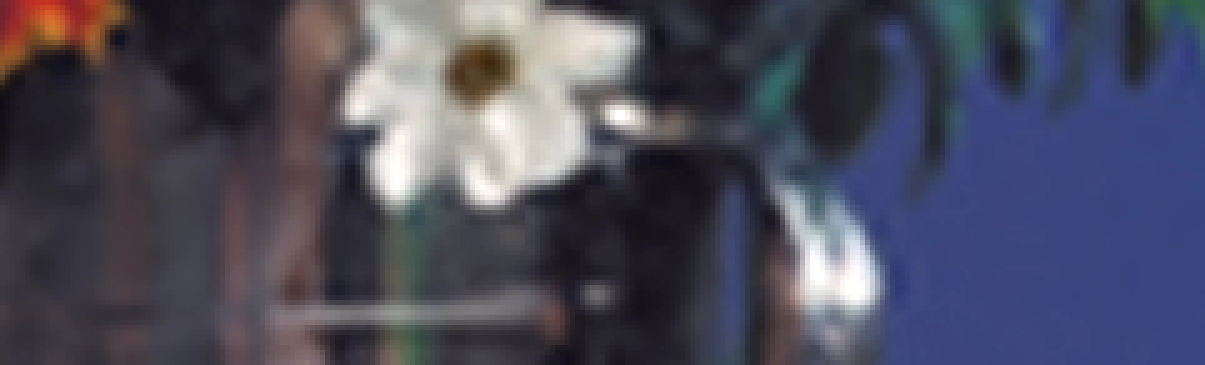}&
\includegraphics[width=0.3\linewidth]{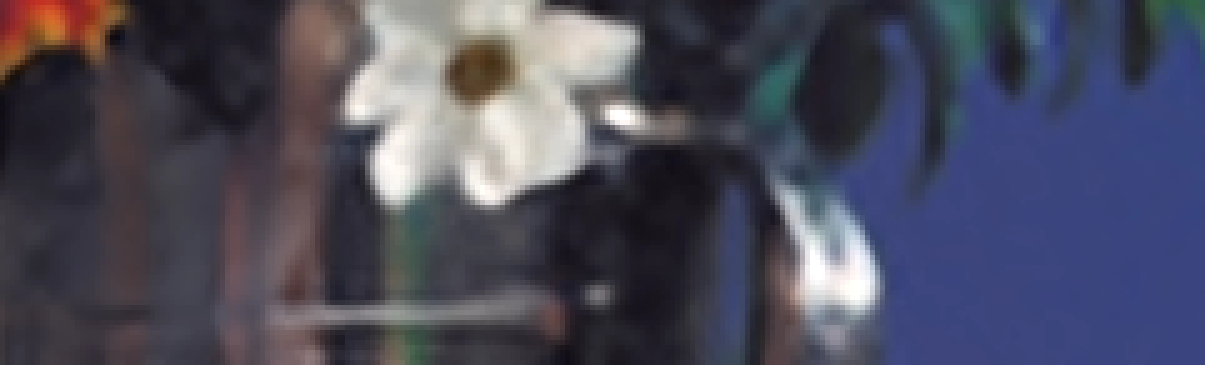}&
\includegraphics[width=0.3\linewidth]{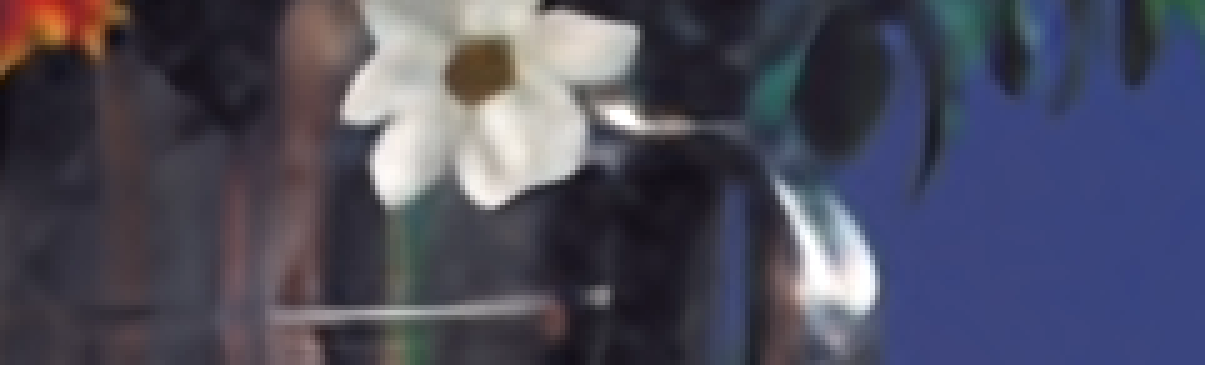}\\
\includegraphics[width=0.075\linewidth]{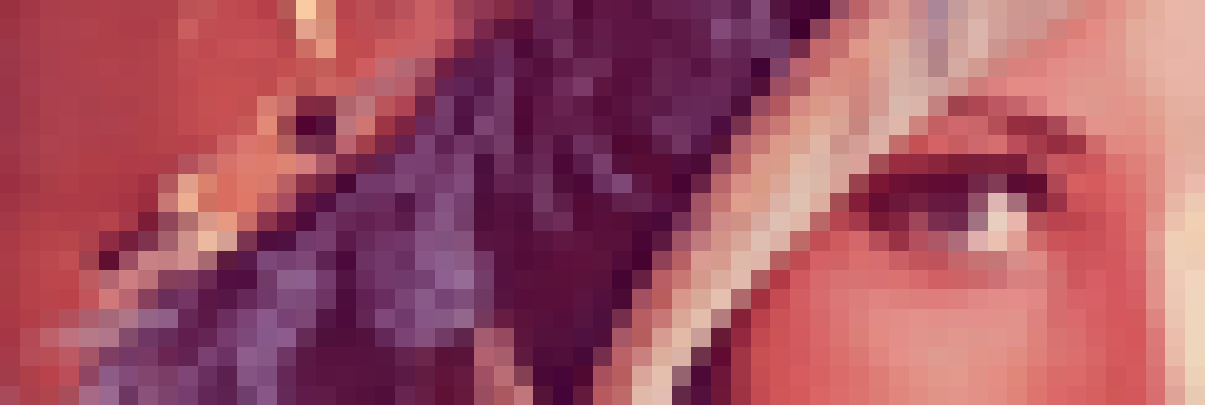}&
\includegraphics[width=0.3\linewidth]{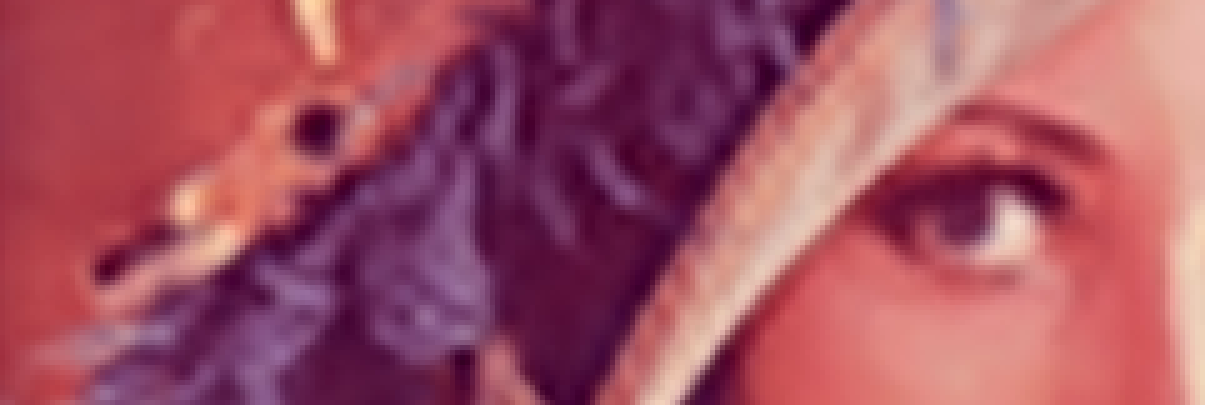}&
\includegraphics[width=0.3\linewidth]{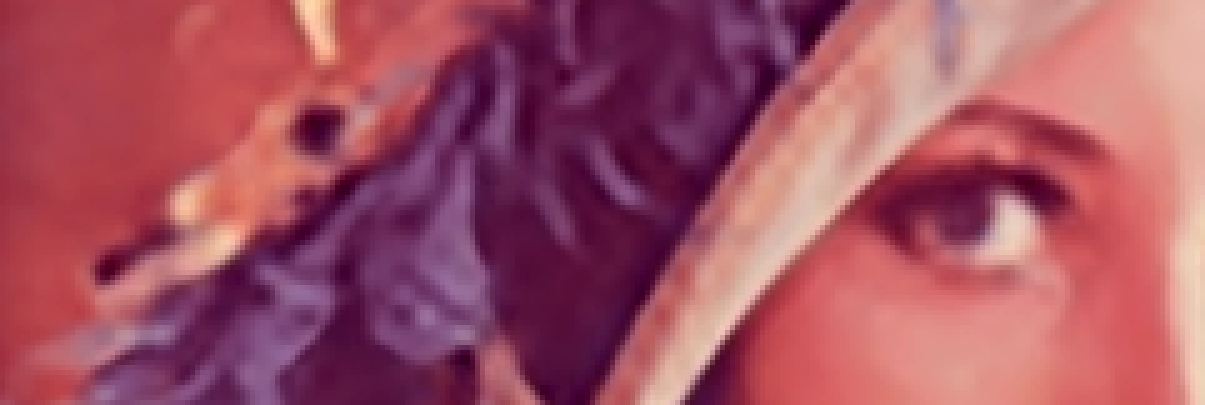}&
\includegraphics[width=0.3\linewidth]{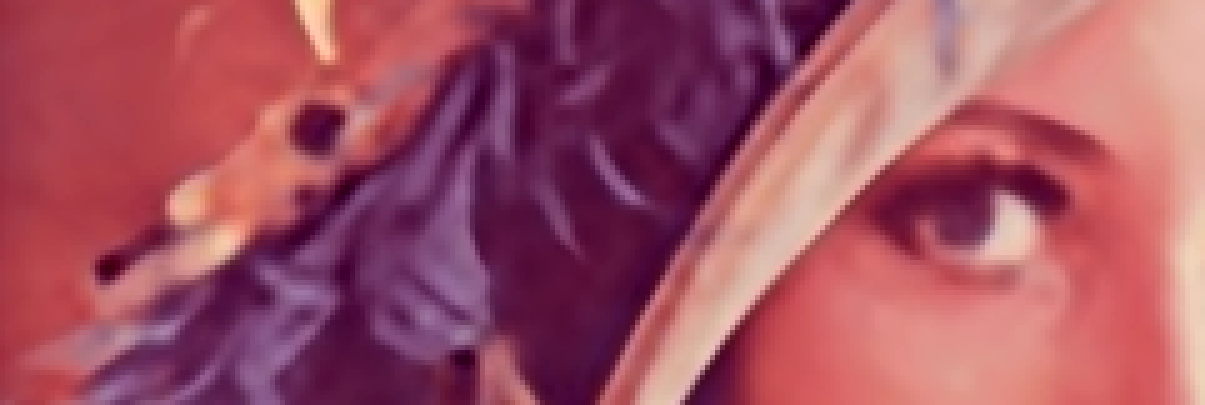}\\
\includegraphics[width=0.075\linewidth]{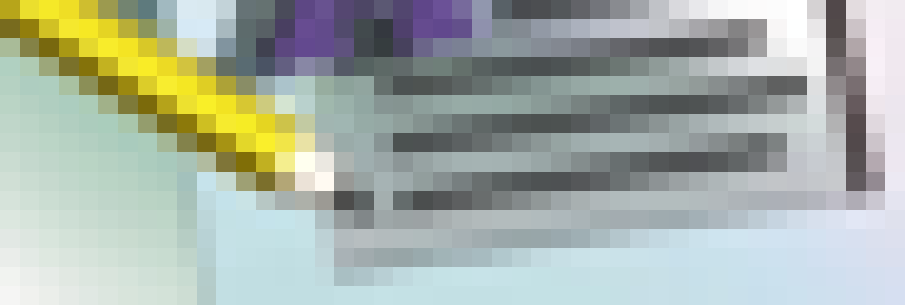}&
\includegraphics[width=0.3\linewidth]{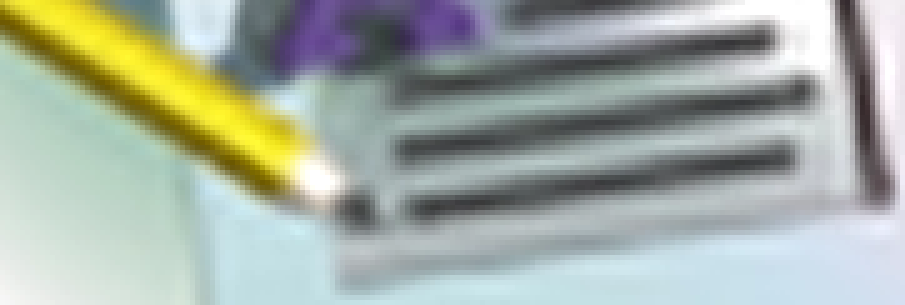}&
\includegraphics[width=0.3\linewidth]{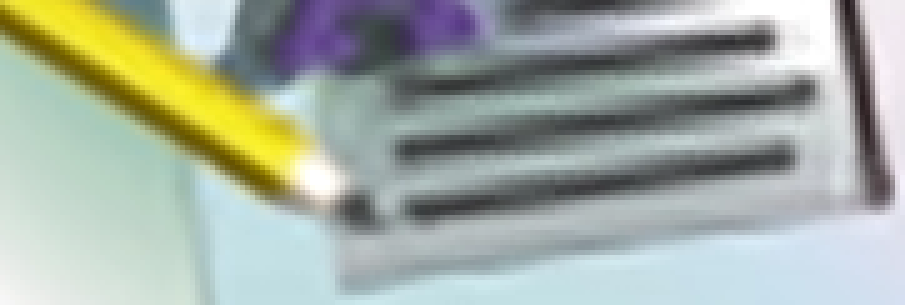}&
\includegraphics[width=0.3\linewidth]{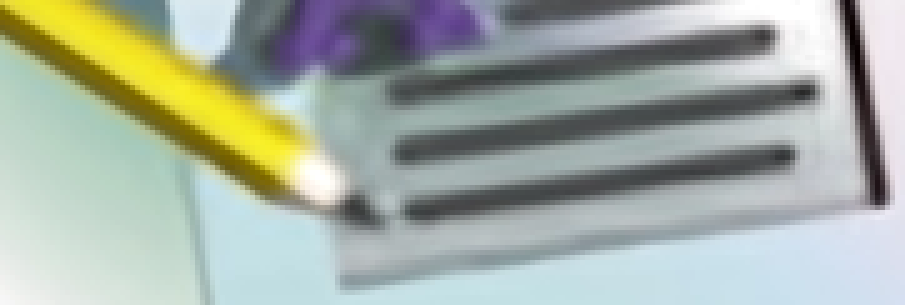}\\
\includegraphics[width=0.1\linewidth]{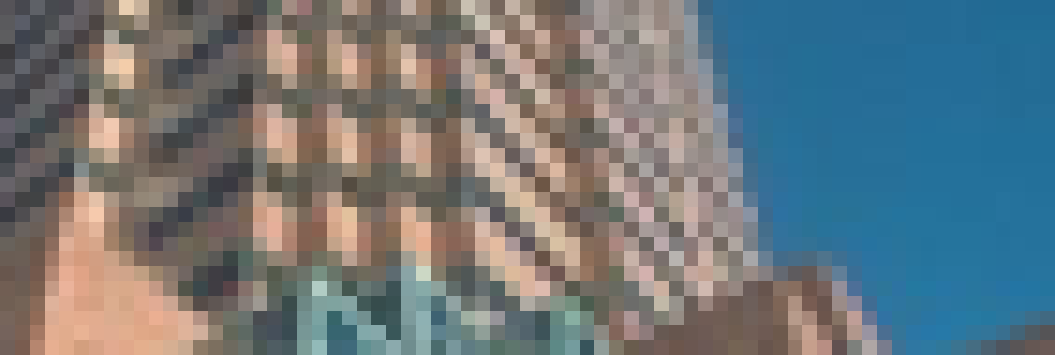}&
\includegraphics[width=0.3\linewidth]{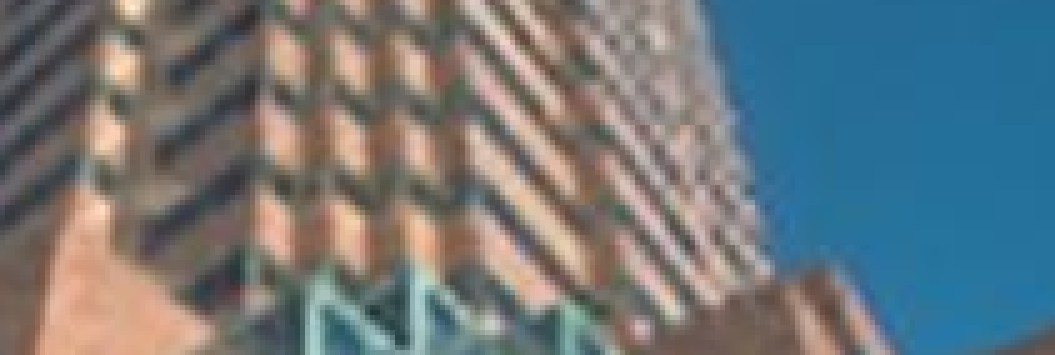}&
\includegraphics[width=0.3\linewidth]{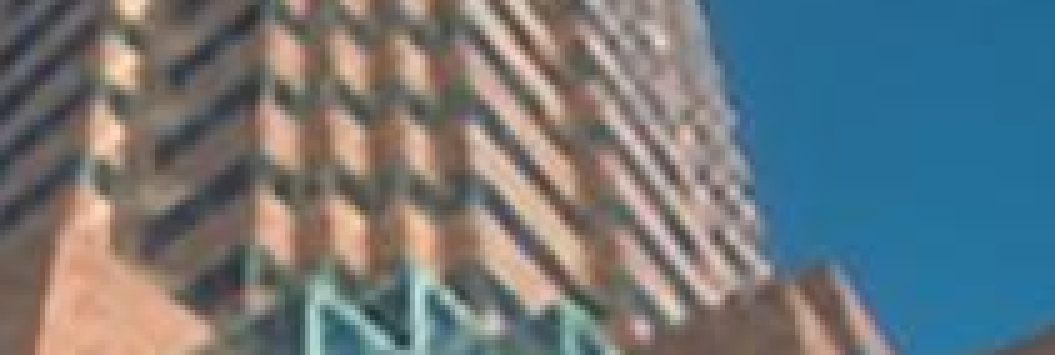}&
\includegraphics[width=0.3\linewidth]{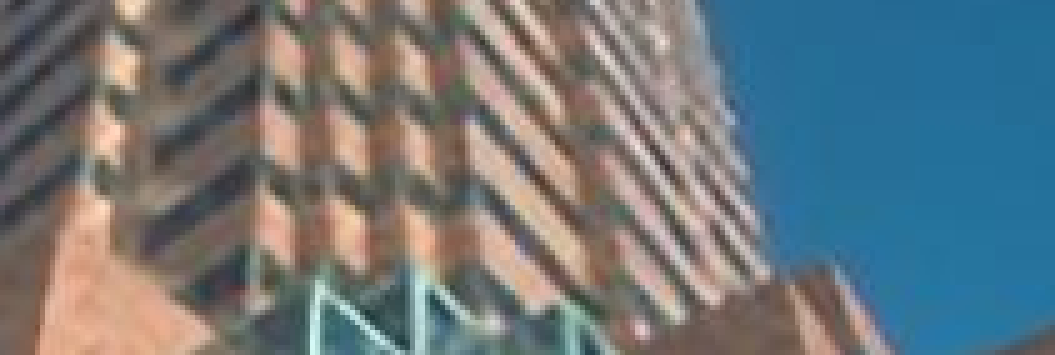}\\
\includegraphics[width=0.1\linewidth]{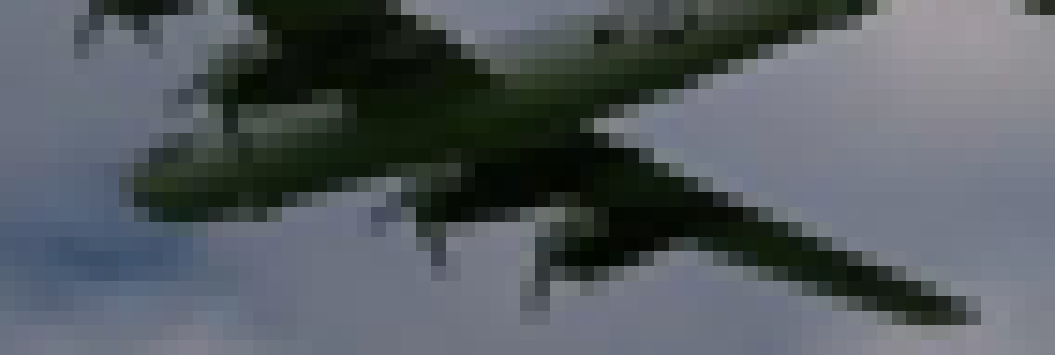}&
\includegraphics[width=0.3\linewidth]{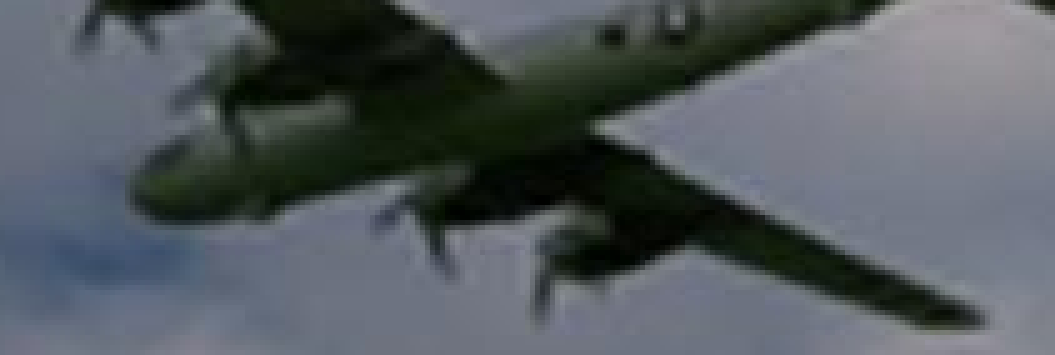}&
\includegraphics[width=0.3\linewidth]{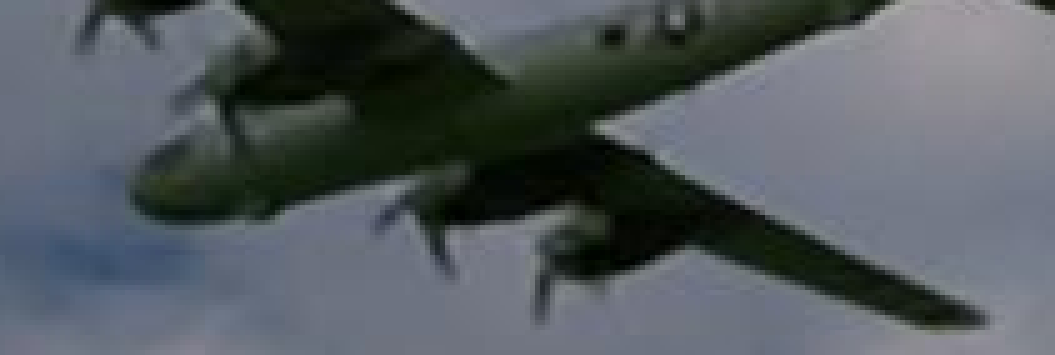}&
\includegraphics[width=0.3\linewidth]{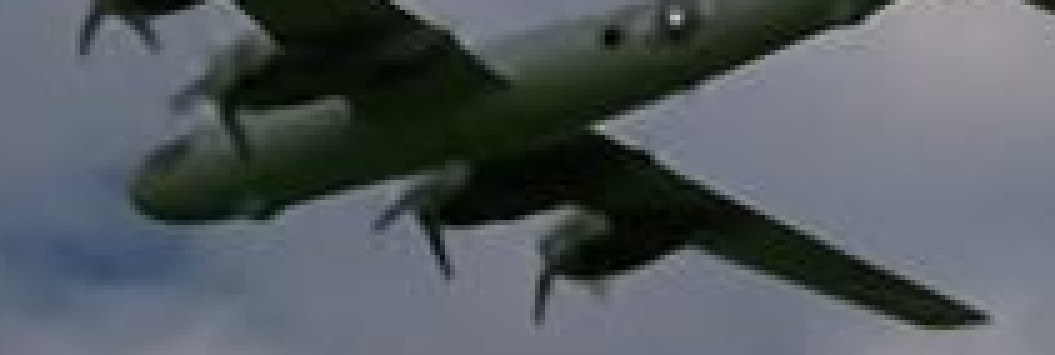}
\end{tabular}
}
\caption{SR visual comparison. \textbf{Best zoomed in on screen.}}
\label{fig:visual_crops}
\vspace{-0.5cm}
\end{figure}

\begin{figure*}[]
\centering
\setlength{\tabcolsep}{1pt}
\renewcommand{\arraystretch}{0.2} 
\resizebox{\linewidth}{!}
{
\large
\begin{tabular}{ccccc}
\parbox[b]{3mm}{\rotatebox[origin=l]{90}{Bicubic}}&
\includegraphics[height=5cm]{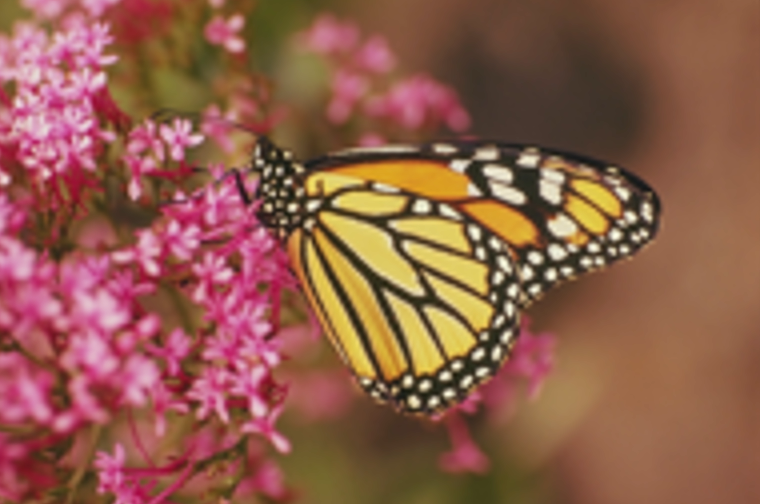}&
\includegraphics[height=5cm]{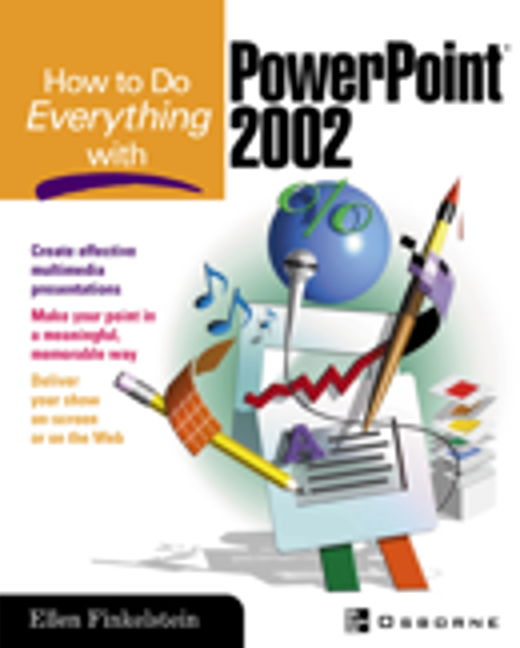}&
\includegraphics[height=5cm]{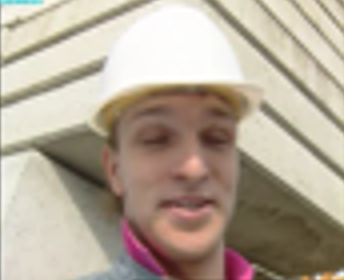}&
\includegraphics[height=5cm]{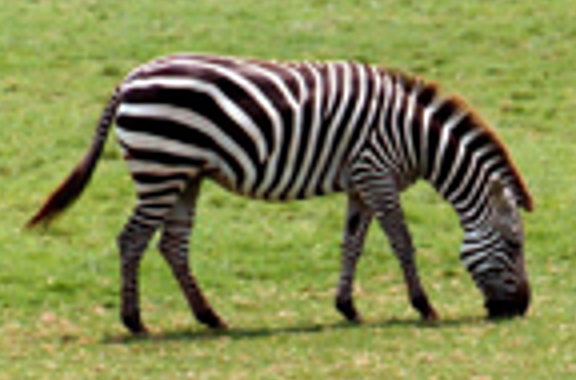}\\[4pt]
\parbox[b]{3mm}{\rotatebox[origin=l]{90}{Zeyde~\cite{Zeyde-CS-2012}}}&
\includegraphics[height=5cm]{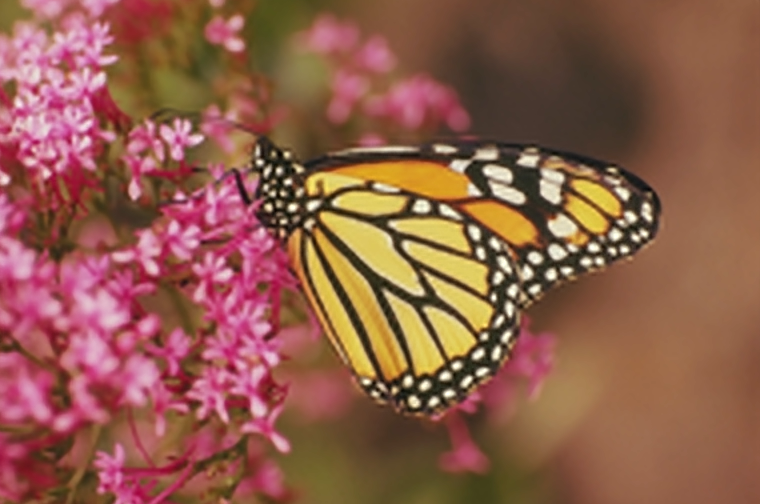}&
\includegraphics[height=5cm]{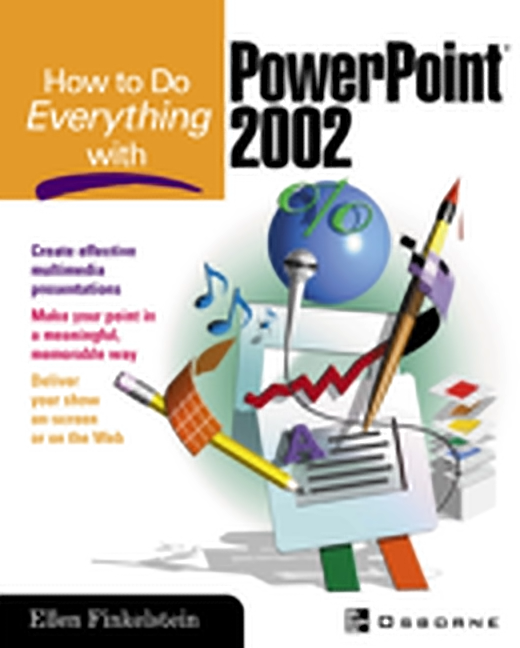}&
\includegraphics[height=5cm]{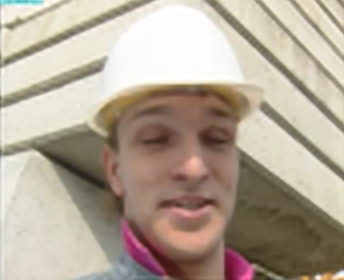}&
\includegraphics[height=5cm]{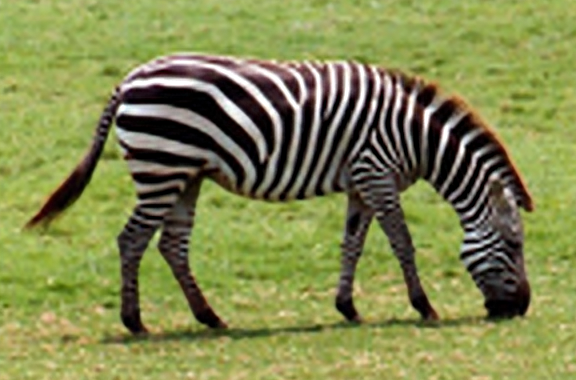}\\[4pt]
\parbox[b]{3mm}{\rotatebox[origin=l]{90}{ANR~\cite{Timofte-ICCV-2013}}}&
\includegraphics[height=5cm]{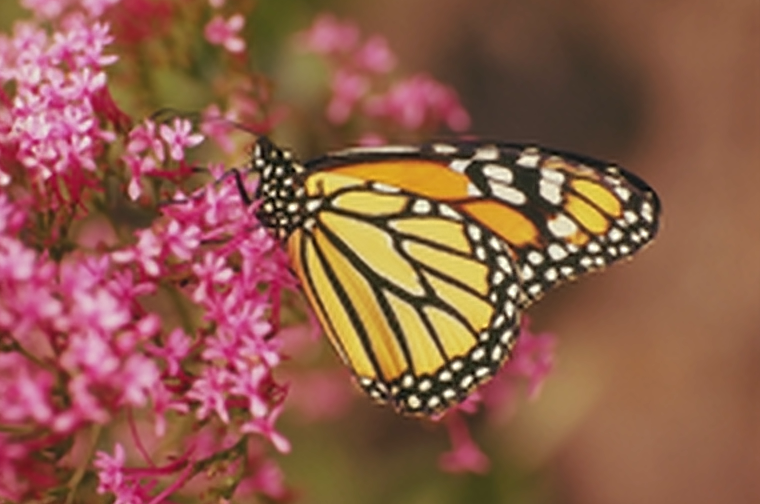}&
\includegraphics[height=5cm]{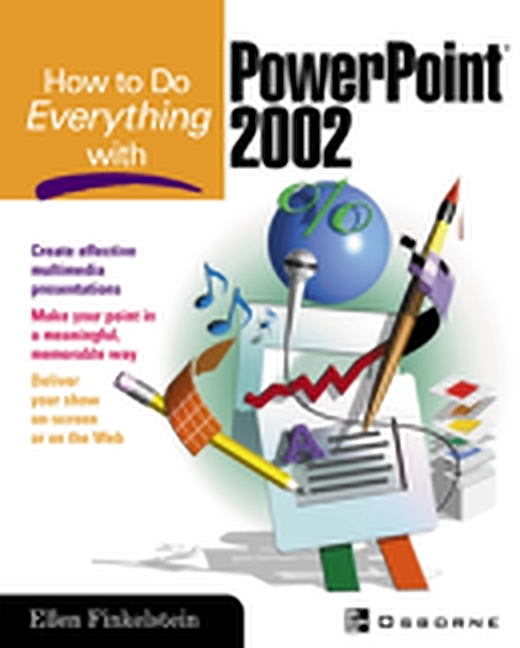}&
\includegraphics[height=5cm]{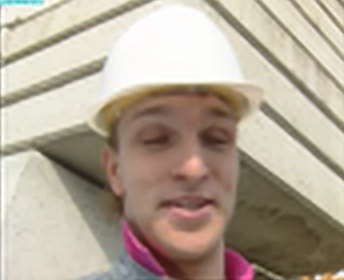}&
\includegraphics[height=5cm]{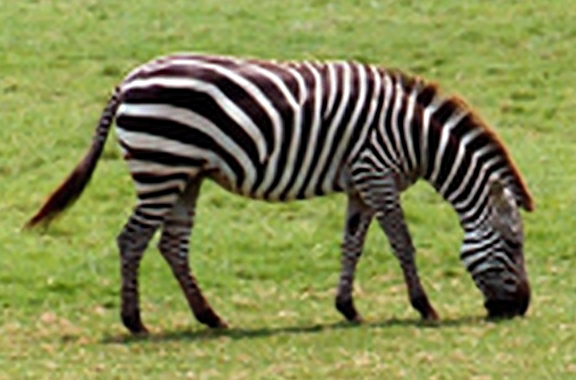}\\[4pt]
\parbox[b]{3mm}{\rotatebox[origin=l]{90}{A+~\cite{Timofte-ACCV-2014}}}&
\includegraphics[height=5cm]{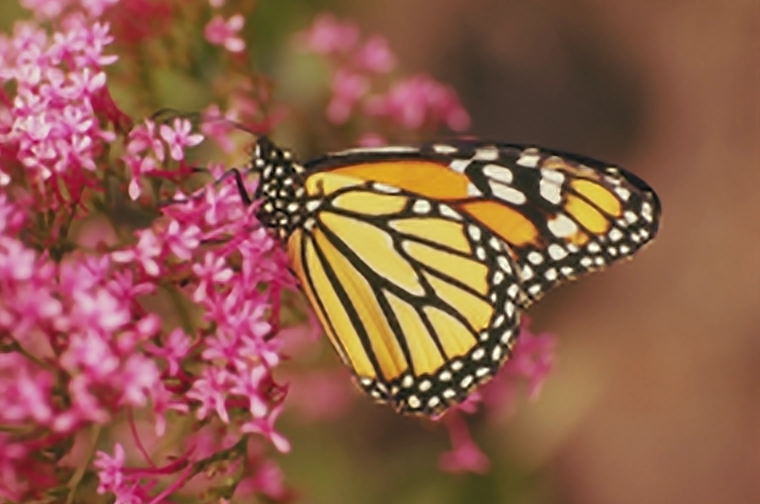}&
\includegraphics[height=5cm]{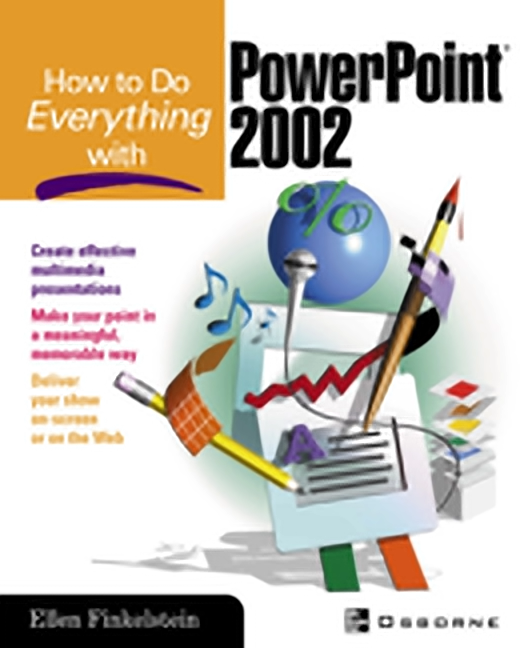}&
\includegraphics[height=5cm]{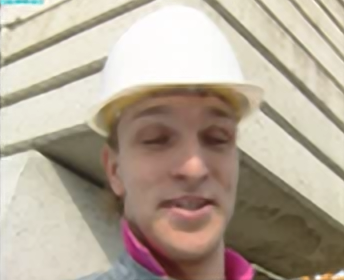}&
\includegraphics[height=5cm]{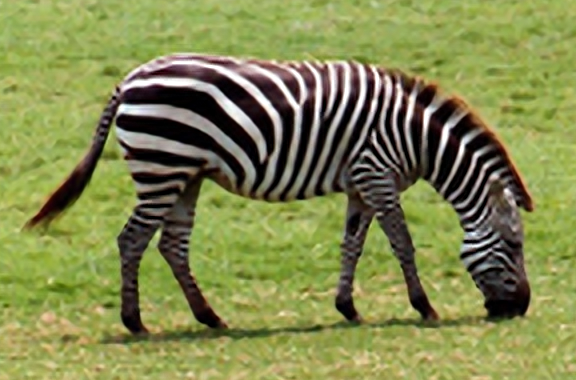}\\[4pt]
\parbox[b]{3mm}{\rotatebox[origin=l]{90}{{\bf IA (ours)}}}&
\includegraphics[height=5cm]{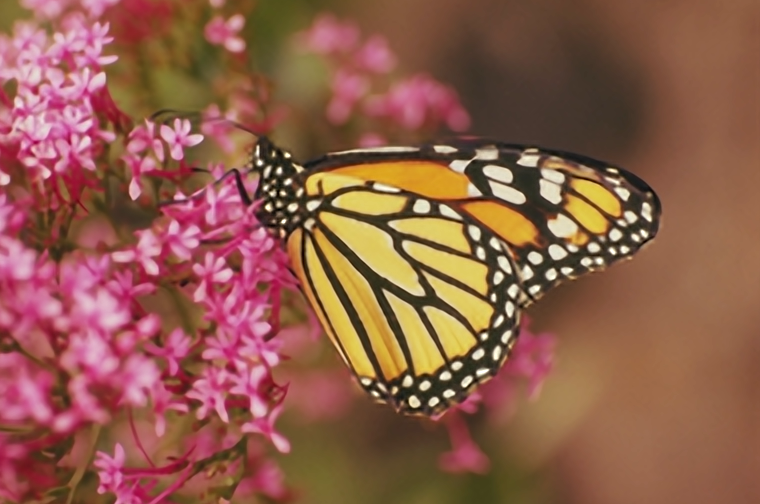}&
\includegraphics[height=5cm]{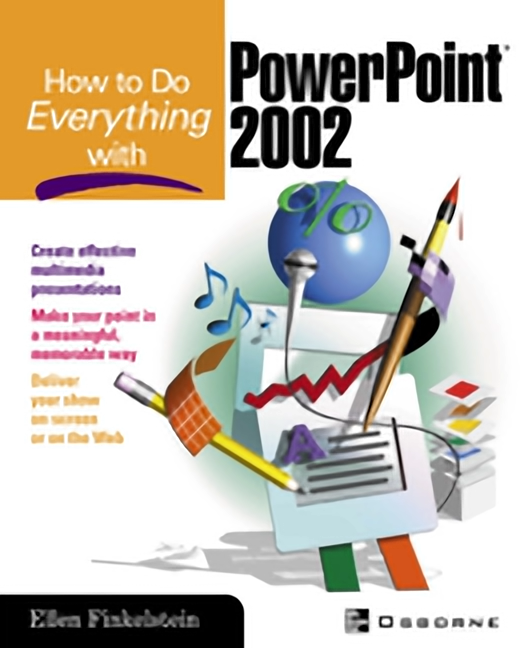}&
\includegraphics[height=5cm]{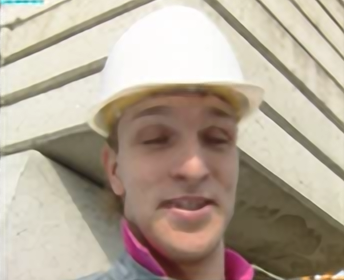}&
\includegraphics[height=5cm]{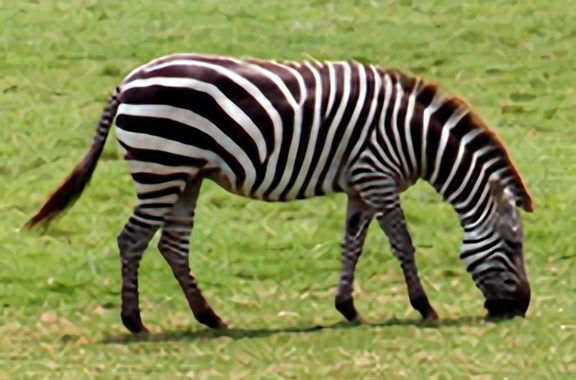}\\[4pt]
\end{tabular}
}
\caption{SR visual results for $\times4$. Images from Set14. \textbf{Best zoomed in on screen.}}
\label{fig:visual_results}
\vspace{0.5cm}
\end{figure*}

For qualitatively assessing the IA performance we depict in Fig.~\ref{fig:visual_results} several cropped images for magnification factors $\times3$ and $\times4$. Generally IA restores more sharp details with fewer artifacts than the A+ and Zeyde methods do. For example, the clarity and sharpness of the HR reconstruction for the text image visibly improves from the Zeyde to A+ and then to our IA result.
The same for other face features such as chin, mouth or eyes in the image from the second row of Fig.~\ref{fig:visual_crops}.

In Fig.~\ref{fig:visual_results} we show image results for magnification $\times4$ on Set14 for our IA method in comparison with the bicubic, Zeyde, ANR, and A+ methods.

The supplementary material contains more per image PSNR results and HR outputs for qualitative assessment.

\section{Conclusion}
\label{sec:conclusion}

We proposed seven ways to effectively improve the performance of example-based super-resolution. Combined, we obtain a new highly efficient method, called Improved A+ (IA), based on the anchored regressors idea of A+.
Non-invasive techniques such as augmentation of the training data, enhanced prediction by consistency checks, context reasoning, or iterative back projection lead to a significant boost in PSNR performance without significant increases in running time.
Our hierarchical organization of the anchors in the IA method allows us to handle orders of magnitude more regressors than the original A+ at the same running time.
Another technique, often overlooked, is the cascaded application of the core super-resolution method towards HR restoration. Using the image self-similarities or the context is shown also to improve PSNR.
On standard benchmarks IA improves 0.4dB up to 0.9dB over state-of-the-art methods such as A+~\cite{Timofte-ACCV-2014} and SRCNN~\cite{Dong-ECCV-2014}. 
While we demonstrated the large improvements mainly on the A+ framework, and several other methods (ANR, Yang, Zeyde, SRCNN), we strongly believe that the proposed techniques provide similar benefits for other example-based super-resolution methods. The proposed techniques are generic and require no changes to the core baseline method. 

\newpage
{
\small
\bibliographystyle{ieee}
\bibliography{improvedSISR_arXiv}
}

\end{document}